\documentclass{article}
\usepackage{log_2025}				

\usepackage{booktabs}						
\usepackage{multirow}						
\usepackage{amsfonts}						
\usepackage{graphicx}						
\usepackage{duckuments}						
\usepackage[utf8]{inputenc} 
\usepackage[T1]{fontenc}    
\usepackage{url}            
\usepackage{booktabs}       
\usepackage{amsfonts}       
\usepackage{nicefrac}       
\usepackage{microtype}      
\usepackage{xcolor}         

\usepackage{amsmath}
\usepackage{amssymb}
\usepackage{mathtools}
\usepackage{amsthm}
\usepackage{subcaption}
\usepackage{wrapfig}

\theoremstyle{plain}
\newtheorem{theorem}{Theorem}[section]

\newtheorem{lemma}[theorem]{Lemma}

\theoremstyle{definition}
\newtheorem{definition}[theorem]{Definition}

\theoremstyle{remark}

\usepackage[textsize=tiny]{todonotes}

\usepackage{colortbl}
\usepackage{microtype}
\usepackage{graphicx}
\usepackage{comment}
\usepackage{booktabs} 
\usepackage{threeparttable} 
\usepackage{comment}
\newcommand{\R}{\mathbb{R}}
\usepackage[numbers,compress,sort]{natbib}	

\title[Position: Beyond Euclidean -- Foundation Models Should Embrace Non-Euclidean Geometries]{Position: Beyond Euclidean -- Foundation Models Should Embrace Non-Euclidean Geometries}

\author[He et al.]{%
Neil He\thanks{Equal contribution}\\
Yale University \\
\email{neil.he@yale.com}
\And
Jiahong Liu\footnotemark[1]\\
Chinese University of Hong Kong\\
\email{jiahong.liu21@gmail.com}
\And
Buze Zhang\\
Xi’an Jiaotong
University\\
\email{buzejin2022@gmail.com}
\And
Ngoc Bui\\ Yale University \\
\email{ngoc.bui@yale.edu}
\And
Ali Maatouk \\ Yale University \\
\email{ali.maatouk@yale.edu}
\And
Menglin Yang\thanks{Corresponding author, work done while at Yale University} \\ Hong Kong University of Science and Technology \\
\email{menglin.yang@outlook.com}
\And 
Irwin King \\ Chinese University of Hong Kong\\
\email{king@cse.cuhk.edu.hk}
\And 
Melanie Weber \\ Harvard University \\
\email{mweber@g.harvard.edu}
\And 
Rex Ying \\ Yale University \\
\email{rex.ying@yale.edu}
}
\usepackage[backref=page]{hyperref}
\usepackage[capitalize,noabbrev]{cleveref}

\begin{document}

\maketitle

\begin{abstract}
In the era of foundation models and Large Language Models (LLMs), Euclidean space has been the de facto geometric setting for machine learning architectures. However, recent literature has demonstrated that this choice comes with fundamental limitations. At a large scale, real-world data often exhibits inherently non-Euclidean structures, such as multi-way relationships, hierarchies, symmetries, and non-isotropic scaling, in a variety of domains, such as languages, vision, and the natural sciences. 
It is challenging to effectively capture these structures within the constraints of Euclidean spaces.
This position paper argues that moving beyond Euclidean geometry is not merely an optional enhancement but a necessity to maintain the scaling law for the next-generation of foundation models. 
By adopting these geometries, foundation models could more efficiently leverage the aforementioned structures. Task-aware adaptability that dynamically reconfigures embeddings to match the geometry of downstream applications could further enhance efficiency and expressivity. 
Our position is supported by a series of theoretical and empirical investigations of prevalent foundation models.
Finally, we outline a roadmap for integrating non-Euclidean geometries into foundation models, including strategies for building geometric foundation models via fine-tuning, training from scratch, and hybrid approaches.
\end{abstract}

\section{Introduction}
\begin{wrapfigure}{r}{0.4\textwidth}
\vspace{-37pt}
\centering
    \includegraphics[width=0.35\textwidth]{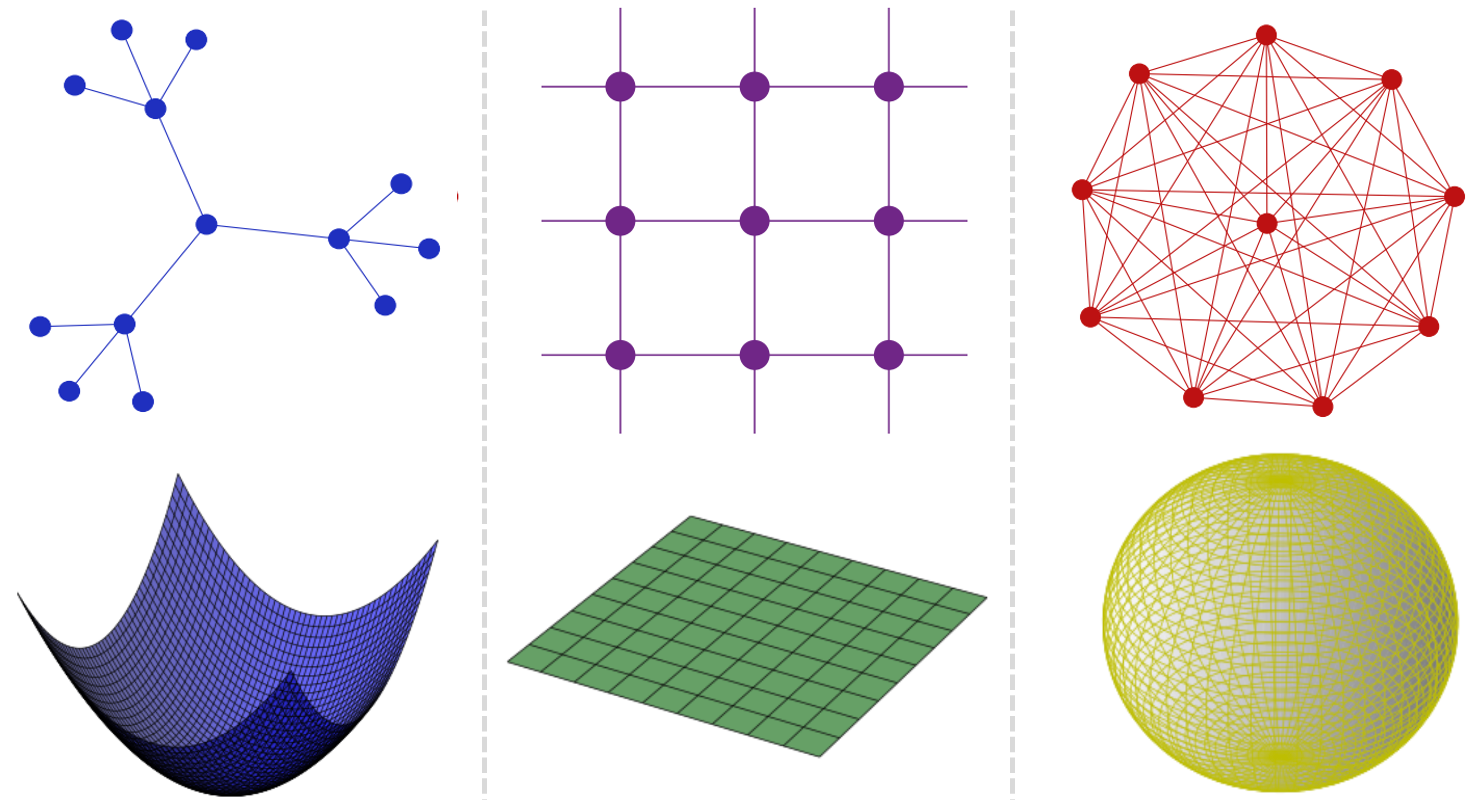}
    \caption{\small Manifolds and their corresponding graph structures or underlying relationships, which represent different types of token relationships: hierarchical (left), uniform (middle), and cyclical (right) dependencies.
}
    \label{figure:graph_geometry}
    \vspace{-22pt}
\end{wrapfigure}
Foundation models, such as Large Language Models (LLMs), have emerged as a cornerstone of current AI advancements due to their ability to generalize across diverse tasks with minimal fine-tuning~\cite{bommasani2021on,devlin2018bert,brown2020language,ramesh2021DALLE}. Euclidean geometry has been the default framework for designing such models, largely driven by the natural compatibility of Euclidean geometry with fundamental neural network operations—such as linear transformations, convolutions, and attention mechanisms—which can be executed efficiently using standard linear algebra in Euclidean space. While efficient and intuitive, this choice of Euclidean geometry introduces biases and distortions as a result of its flat geometry and linear assumption, such as in the distances computed for downstream tasks. However, \textbf{real-world datasets often exhibit implicit non-Euclidean structures}, such as the hierarchical organization of natural language—including concept taxonomies and entailment relationships~\cite{nickel2017poincare, tifrea2019poincare, le2019inferring}—as well as hierarchical relationships among object classes, scenes, and their constituent categories in visual data~\citep{ge2023hyperbolic, pal2024compositional}. Furthermore, non-Euclidean characteristics are inherent in biological data, such as protein structures \cite{villegas2021foldhsphere} and RNA-seq data \cite{klimovskaia2020poincare}. Given the non-Euclidean characteristics of training data, along with the challenges faced by current foundation models—from hallucinations to computational inefficiencies—it \textbf{becomes crucial to question whether Euclidean geometry should remain the default for foundation models.} Existing works that attempt to incorporate non-Euclidean geometry into neural networks focus on finding geometry alternatives to fundamental Euclidean neural network operations that achieves similar goals as the Euclidean counterparts, such as defining neural network layers in non-Euclidean embedding spaces~\cite{chen2021fully, yang2024hypformer, HNN++, he2025lresnet} or preserves certain geometric properties~\cite{lai2023spherical, cohen2018spherical, deepsphere2020deepsphere} or designing loss functions in a non-Euclidean latent space~\cite{ge2023hyperbolic, desai2023hyperbolic}. However, the majority of these works do not actively attempt to address the problem at the level of foundation models and instead focus on low-dimensional settings. See \cref{extended_related_work} for further discussion of non-Euclidean geometry and the geometric assumptions of embedding spaces. 

\textbf{Position: The development of non-Euclidean foundation models is essential for effectively representing, modeling, and analyzing complex data structures and relationships in real-world applications.} Particularly, this paper advocates for the development of \textit{\textbf{non-Euclidean foundation models}} at the scale of billions of parameters, which is on both a much larger scale and a much broader geometric scope than existing research that focuses almost entirely on low-dimensional settings within specific geometries, such as the hyperbolic space. Such research requires \textbf{significant community efforts}. Beyond proposing non-Euclidean counterparts of Euclidean operations, which is the current focus, we advocate building the full training infrastructure for this scale, developing and training large-scale non-Euclidean architectures, and analyzing their behavior, which often diverges from low-dimensional models. 
With arguments grounded in theoretical insights and experimental evidence, we contend that by aligning foundation models—whether visual, linguistic, or scientific—with the intrinsic geometries of their training data, we can improve three critical aspects of these models: representational capabilities, adaptability to diverse geometric structures, and scalability.

\textbf{Representational Capabilities}. Euclidean space has limited capabilities to represent complex geometric structures with diverse local properties, resulting in significant distortion when embedding such data in low-dimensional Euclidean spaces~\cite{tenenbaum2000isomap}. In contrast, hyperbolic spaces, with their negative curvature, excel at representing hierarchical structures with minimal distortion in low dimensions~\cite{krioukov2010hyperbolic,sarkar2011low}. Similarly, spherical geometries, defined by positive curvature, are well-suited for modeling data with bounded structures and angular relationships~\cite{durrant2022hyperspherically,esteves2017learning,trosten2023hubs}.

\textbf{Adaptability.} 
Incorporating non-Euclidean geometrical operations into foundation models provides substantially enhanced adaptability to the diverse geometric structures in training datasets, particularly in the case of large-scale datasets---as is typical for these models—where heterogeneity is inherent by design.
This adaptability 
improves the models' flexibility and transferability, as many instances of downstream tasks benefit from acknowledging the geometric structure in the data, such as knowledge graph completion~\cite{bai2021modeling,wang2021mixed}, social network analysis~\cite{yang2022hyperbolic,kennedy2013hyperbolicity}, multi-label classification, drug discovery~\cite{poleksic2023hyperbolic}, user preferences recommendation~\cite{chen2022modeling,yang2022hrcf,yang2022hicf}, and code understanding~\cite{tang2023hyperbolic}. 

\textbf{Scalability.} Adapting foundation models to non-Euclidean geometry enables expressive lower-dimensional embeddings, reducing computational costs without sacrificing performance. A critical implication lies in the scaling laws of foundation models~\cite{10.5555/3600270.3602446}, where performance in Euclidean models follows a power-law scaling of the form $L(N) \propto N^{-\alpha}$, with $L$ and $N$ being the loss and parameter count. This behavior reveals inherent inefficiencies in how Euclidean space handles increasing model complexity and data dimensionality. In contrast, Riemannian methods have shown promises to improve scaling by efficiently compressing information~\cite{brehmer2024does,kiani2024hardness}. For instance, hyperbolic spaces better captures long-range dependencies~\cite{tifrea2019poincare} and mixed-curvature approaches~\cite{gu2019mixedcurvature,wang2021mixed} allow different model components to scale according to their optimal geometric properties.

\textbf{Roadmap.} Moreover, we propose a roadmap for integrating non-Euclidean geometries into foundation models. This includes both adapting existing Euclidean models to incorporate these principles and developing foundation models from scratch. We also highlight key challenges and outline the steps required to advance this non-Euclidean vision, from architectural design to the creation of non-Euclidean libraries, given that existing frameworks such as DeepSpeed~\cite{rasley2020deepspeed} and Flash Attention~\cite{dao2022flashattention} are tailored exclusively for Euclidean models. 

\section{Background and Preliminaries}
In this section, we give an overview of non-Euclidean spaces, particularly focusing on Riemannian manifolds. For more details please see~\cite{lee2013manifold} and \cref{extended_related_work}.
\subsection{Non-Euclidean Geometry Foundations}
\textbf{Riemannian Manifolds.} A smooth $n$-dimensional manifold $\mathcal{M}$ is a topological space that is locally Euclidean. Each point $x$ is associated with a \textit{tangent space} $T_x\mathcal{M}$, which is an $n$-dimensional vector space that acts as a first-order local approximation of $\mathcal{M}$. A \textit{Riemannian metric} $\mathfrak{g}$ on $\mathcal{M}$ is a collection $\mathfrak{g}\coloneqq(\mathfrak{g}_x)_{x\in\mathcal{M}}$ of positive definite bilinear forms $\mathfrak{g}_x(\cdot, \cdot): T_x\mathcal{M}\times T_x\mathcal{M}\to \R^n$, varying smoothly with $x$. $\mathfrak{g}_x$ induces the \textit{(sectional) curvature} at point $x$, which measures how $\mathcal{M}$ deviates from flatness at $x$. A \textit{Riemannian manifold} is a pairing $(\mathcal{M}, \mathfrak{g})$. For example, $\R^n$ with the usual Euclidean inner product is a Riemannian manifold with constant curvature $0$. $\mathfrak{g}_x$ can be seen as a generalization of inner products, where the norm of $p\in T_x\mathcal{M}$ is $\|p\|_\mathfrak{g} = \sqrt{\mathfrak{g}_x(p, p)}$. The choice of $\mathfrak{g}$ induces a global distance function $d(\cdot, \cdot)$ on $\mathcal{M}$. A \textit{geodesic} between $x,y$ is a local distance minimizing smooth curve. In particular, the shortest paths are geodesics. With certain assumption on the structure of $\mathcal{M}$, one can define the \textit{exponential map} $\exp_x:T_x\mathcal{M}\to \mathcal{M}$ for $x\in\mathcal{M}$, and its inverse, the \textit{logarithmic map} $\log_x:\mathcal{M}\to T_x\mathcal{M}$. Additionally, the \textit{parallel transport} map $\mathrm{PT}_{x}(v, w)$, where $v,w\in T_x\mathcal{M}$, generalizes  translation, transporting $w$ starting at $x$ in the direction of $v$ with no acceleration.

\subsection{Deep Learning in Non-Euclidean Spaces}
Recent years have witnessed an increasing interest in extending deep learning techniques to Riemannian manifolds. We discuss several advances for designing neural networks and Transformers in non-Euclidean geometries, as well as optimization on manifolds, with more details in \cref{extended_related_work}.

\textbf{Non-Euclidean Neural Networks.} Several works have explored neural networks that leverage geodesic distance to perform neural network operations~\cite{bronstein2017geometric, masci2015geodesic, boscaini2016learning, kong2022geodesic}. Within hyperbolic learning, prior works have developed neural network layers~\cite{HNN, HNN++, nickel2017poincare, chen2021fully,weber2020robust}, graph neural networks~\cite{liu2019HGNN, hgcn2019}, vision models~\cite{Bdeir2024fully, van2023poincar}, and residual neural networks~\cite{he2025lresnet}. In addition, extensive works have developed equivariant neural networks that encode spherical geometry as inductive bias~\cite{cohen2018spherical, esteves2017learning, coors2018spherenet, deepsphere2020deepsphere, esteves2019equaivariant}. Neural networks for mixed curvature manifolds that encompass both hyperbolic and spherical models have also been proposed~\cite{gu2019mixedcurvature, bachmann2020constant}. Many Euclidean convex and stochastic optimization algorithms have been extended to manifold learning as well~\cite{udriste1994convex, zhang2016riemannian, becigneul2018riemannian, weber2021projection, weber2022riemannian}.

\textbf{Non-Euclidean Transformers.} Significant advancements have been made toward Transformers in non-Euclidean spaces in
recent studies. Prior works have developed attention mechanisms and additional essential operations, such as layer normalization, to develop Transformers in hyperbolic, spherical, and mixed curvature manifolds~\cite{gulcehre2019hyperbolicAT, chen2021fully, HNN++, yang2024hypformer, lai2023spherical, cho2023curve}.

Nevertheless, there is \textbf{a lack of works for non-Euclidean foundation models}. While prior works have attempted or argued for incorporating geometric or topological information into certain aspects of model design~\cite{papillon2025beyondeuclid, bronstein2017geometric, hajij2022topological}, these prior works almost all focus on low-dimensional settings, with few works that consider pre-trained models~\cite{chen2024hypbert}, thus omitting the vast potential benefits to be gained from non-Euclidean foundation models.

\section{Foundation Models Should Embrace Non-Euclidean
Geometries}

\textbf{Euclidean Foundation Models.} 
Foundation models are typically trained on massive corpora to learn transferable representations that serve as a basis for downstream tasks \cite{bommasani2021on}. Transformer-based language models~\cite{devlin2018bert, brown2020language, brown2020language, 2020t5, dubey2024llama3}, large-scale vision models such as Vision Transformer (ViT) and ResNet \cite{dosovitskiy2020image, he2016deep}, and multimodal foundation models like CLIP \cite{radford2021CLIP} and DALL-E \cite{ramesh2021DALLE}, have achieve state-of-the-art performances in a vast amount of tasks across numerous domains.

\subsection{Limitations of Euclidean Geometry for Foundation Models}\label{sec:euclidean_limitation}

The Euclidean assumption is that relationships between data points can be meaningfully characterized using distances measured in a flat space. However, theoretical and experimental works have demonstrated that \textbf{Euclidean geometry, with its isotropic nature and uniform scaling, fails to capture the complex structures of real-world data, resulting in significant distortions}~\citep{bourgain1986metrical, matouvsek1996distortion, arias1992finite, gupta1999embedding, matouvsek1999embedding}. As a result, high-quality, low-distortion embeddings are often only possible in \emph{high-dimensional} Euclidean space. Specifically, embeddings of complex structured data, such as hierarchies or trees, provably incur \textit{high rates of distortion}~\citep{bourgain1985lipschitz,matouvsek1996distortion}. In this section, we highlight how the flat nature of the Euclidean space results in limitations and challenges for foundational models.

\textbf{Non-Applicability of the Nash Embedding Theorem.} The Nash Embedding Theorem 
states that any Riemannian manifold $\mathcal{M}$ of dimension $n$ admits an isometric embedding $f$ into $\R^{2n+1}$~\cite{nash1954embedding}, seemingly to imply that non-Euclidean spaces would only reduce the embedding dimension by half. However, the isometric embedding here is defined to preserve the Riemannian metric, meaning that it is \textit{locally distance preserving}---the length of any path is preserved. However, for the shortest path between points $x,y\in \mathcal{M}$, its image under $f$ is not necessarily the shortest path (i.e., Euclidean straight line) between $f(x)$ and $f(y)$. Conversely, measuring the embedding distortion is concerned with whether a map is \textit{globally distance preserving} w.r.t the ambient space, or when the shortest path between $x$ and $y$ remains the shortest path between $f(x)$ and $f(y)$, which is defined by isometric embeddings between metric (ambient) spaces. Note that \textbf{an isometric embedding between Riemannian manifolds is in general not an isometric embedding between metric spaces}. 

We are concerned with global distance-preserving embeddings for foundational models, as the distance between any pair of token embeddings is crucial for model training. Thus, \textbf{the Nash Embedding Theorem is not applicable} since global distortion could still arise from isometric embeddings between Riemannian manifolds. For this reason, by ``isometry'', we refer to those between metric spaces. See \cref{nash_embedding} for more details. As the Nash Embedding Theorem is not applicable, Euclidean embeddings suffer from several limitations, which we detail below.

\textbf{Dimensionality.} Euclidean space requires high dimensionality to embed complex structures with low distortion, which ties directly to high demand for model size and computational resources. The following theorem shows the distortion-dimension tradeoff for Euclidean embeddings even in the simple case of unweighted token relationships, in the form of complete graphs.
\begin{theorem}(\citet{matousek2002lectures}) Let $X$ be an $n$-point metric space with uniform distance $1$, i.e., an unweighted complete graph with $n$ nodes. For $\epsilon > 0$, the minimal $d$ such that $X$ can be embedded into $\R^d$ with distortion $(1+\epsilon)$ is $d = \Omega\left(\frac{\log(n)}{\epsilon^2\log(1/\epsilon)}\right)$\label{min_dim}
\end{theorem}
For any $p<2$, $\epsilon^2\log(1/\epsilon)$ tends to $0$ faster than $\epsilon^p$ as $\epsilon\to 0$. As a result, \cref{min_dim} implies that $d$ grows \textit{near-quadratically} w.r.t. inverse distortion. Furthermore, \emph{any} unweighted graph with $n$ nodes can be isometrically embedded into an unweighted complete graph with $n$ nodes. Thus \cref{min_dim} implies the same dimensionality issue for embedding any unweighted graph in Euclidean space.

\textbf{Distortion.} Non-trivial distortion could exist regardless of the dimension of the Euclidean space in the cases of more complex structures. The following theorem implies that a wide range of spaces cannot be isometrically embedded into Euclidean space, based on \textit{Markov convexity} (\cref{sec:markov}).
\begin{theorem}\cite{lee2009trees} Let $(X, d_X), (Y, d_Y)$ be metric spaces. For every $p\in\mathbb{N}$, denote $\Pi_p(X), \Pi_p(Y)$ the \textit{Markov $p$-convexity constant} of $X$ and $Y$ respectively. Let $c_Y(X) = \inf\{\mathrm{dist}(f): f:X\to Y\}$ denote the minimum distortion of embedding $X$ in $Y$. Then $c_Y(X) \geq \frac{\Pi_p(X)}{\Pi_p(Y)}$. \label{dist_theorem}
\end{theorem}
When $X$ models hierarchical token relationships, e.g., $X = B_{2^k}$ is a complete binary tree of depth $2^k$, the distortion for embedding binary trees of depth in any Euclidean space is \textit{at least} $ \Omega(1)\cdot\sqrt{\log k}$. When $X$ represents circular or periodic dependencies in tokens, e.g., $X$
 is a ball of radius $r$ in a vertex-transitive graph, the minimal distortion of embedding $X$ into $\R^n$ for any $n$ is $\Omega(\sqrt{\log r})$~\cite{lee2009trees}.

Moreover, non-trivial distortion exists when embedding other forms of topological space as well, including the sphere $S^k\subseteq \R^{k+1}$, as shown in the following theorem. 
\begin{theorem}\cite{robinson2006sphere}
    Let $(X,d_X)$ be a metric space with $X = \{a, b, c, d\}$ and $d_X(a,b) = d_X(a,c) = d_X(a,d) = 2L$ and $d_X(b,d) = d_X(c,d) = L$ for $L\in\R^+$. Then $X$ admits no isometric embedding into $\R^n$ for any $n$.\label{sphere_dist}
\end{theorem}
As these points can be isometrically embedded into $S^k$, \cref{sphere_dist} shows that $S^k$ cannot be isometrically embedded into $\R^n$ for any $n\in\mathbb{N}$, resulting in distortion when encoding rotational equivariance. In contrast, non-Euclidean geometry can provide a more natural representation of complex topological structures, \textbf{reducing distortion and dimensionality} of the embedding space. For instance, \cite{sarkar2011low} showed that every finite tree admits an embedding into the hyperbolic plane $\mathbb{H}^2$ with $1 + \epsilon$ multiplicative distortion for any $\epsilon > 0$, leading to $O(1)$ distortion with low dimensionality. 

\textbf{Take-away.} 
The implications of the previous theoretical discussion are numerous: (1) \textbf{Limited scalability.} \cref{min_dim} highlights the distortion-dimension trade-off for Euclidean foundation models when embedding complex structures, which is reflected in the computational resources required in these models. Non-Euclidean geometry produces higher quality embeddings in significantly lower dimensions, offering enhanced model scalability; (2) \textbf{Performance bottleneck.} \cref{dist_theorem} and \ref{sphere_dist} demonstrate that even in the case of an abundance of compute resources, the linear assumption in Euclidean foundation models could still incur significant distortion regardless of the embedding dimension for a wide range of topological structures, resulting in a performance upper bound. These theoretical results are validated in \cref{sec:geometry_stats}.

\begin{table*}[]
\centering
\small
\caption{$\delta$-Hyperbolicity of the token embedding in various LLMs across several datasets. The bottom 2 rows show the $\delta$-hyperbolicity values of several metric spaces for reference.}\label{hyperbolicity}
\resizebox{0.98\textwidth}{!}{%
\begin{tabular}{@{}lcccccc@{}}
\toprule
\textbf{Model} & arXiv & C4 & Common Crawl & GitHub & StackExchange & Wikipedia  \\ \midrule
RoBERTa-Base~\cite{liu2019roberta}       & $0.15\pm0.06$               & $0.18\pm0.04$            & $0.17\pm0.04$                      & $0.12\pm0.04$                & $0.17\pm0.07$   & $0.07\pm0.05$  \\
LLaMA3.1-8B~\cite{dubey2024llama3}    & $0.15\pm0.05$               & $0.16\pm0.07$            & $0.15\pm0.06$                      & $0.12\pm0.05$                & $0.18\pm0.06$                       & $0.10\pm0.04$                   \\
GPT-NeoX-20B~\cite{gpt-neox-20b}   & $0.14\pm0.03$               & $0.17\pm0.06$            & $0.15\pm0.05$       & $0.11\pm0.04$                & $0.14\pm0.04$                       & $0.09\pm0.03$                   \\
Gemma2-9B~\cite{gemma2}      &    $0.17\pm0.06$         &      $0.19\pm0.04$                    &        $0.20\pm0.05$                            &              $0.15\pm0.05$               &       $0.18\pm0.04$                              &   $0.15\pm0.03$                              \\ \midrule[0.5pt]

\textbf{Metric Space} & Sphere Space & Dense Graph & PubMed Graph & Poincar{\'e}  Space & Tree Graph & -\\
\textbf{Reference $\delta$ values} & $0.99\pm0.01$ & $0.63\pm0.01$& $0.40\pm0.04$ & $0.14\pm0.01$ & $0.0$ & -\\ \bottomrule
\end{tabular}%
}
\end{table*}

\begin{figure}[!] 
    \centering
       \includegraphics[width=0.99\linewidth]{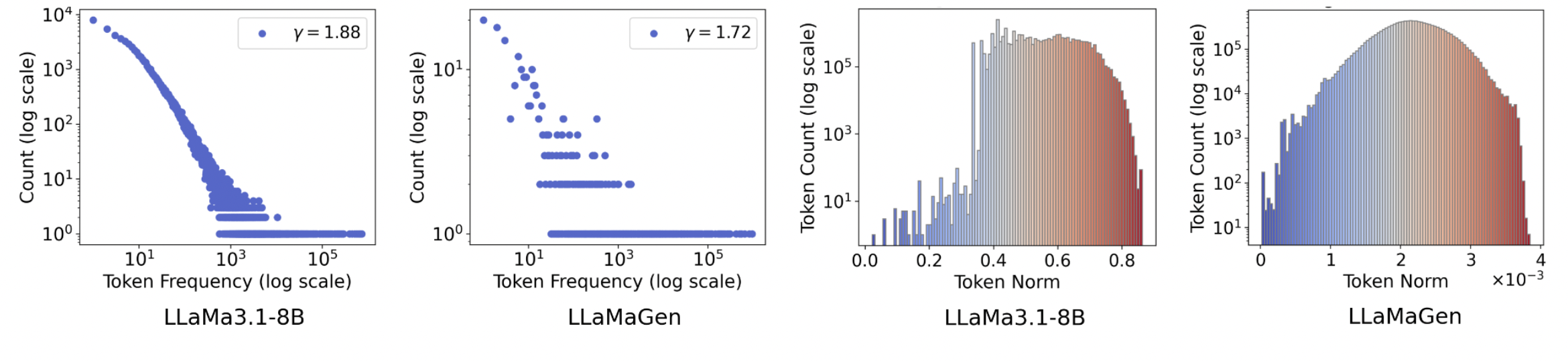}
       \vspace{-5pt}
   \caption{
   Token frequency v.s. token count (left 2) and token norm vs token count (right 2) for LLaMa3.1-8B and LLaMaGen. The datasets are chosen to be within the training corpus. The token-frequency figures show the scale-free properties of the token inputs. The token norm figures reflect this property for learned token embeddings to some extent, with token count increasing exponentially for high-normed tokens at the left tail. However, the Euclidean embeddings still do not fully capture this property and deviate from it at the right tail. 
   More statistics are shown in \cref{additional_statistics}.}
   \label{fig:token_distribution}
   \vspace{-10pt}
\end{figure}

\subsection{Non-Euclidean Geometry in Foundation Models}\label{sec:geometry_stats}

In this section, we empirically assess embedding distortions for different geometries to validate our claims in \cref{sec:euclidean_limitation} and demonstrate that non-Euclidean geometry is more suitable. We then analyze token embeddings in foundation models, showing that \textbf{structures that align with non-Euclidean geometry are prevalent, highlighting the need for alternative geometric frameworks}.

\begin{wraptable}{r}{0.6\textwidth}
  \vspace{-10pt} 
  \begin{minipage}{0.58\textwidth}
    \centering
\centering
\caption{Average (point-wise) distortion on canonical graphs with $96$ nodes, comparing four spaces with total dimension $6$. The least distortion is achieved by the space with the most suitable geometry.}
\resizebox{0.99\textwidth}{!}{%
\begin{tabular}{@{}lccc@{}}
\toprule
\textbf{Geometry}                            & Tree                 & Cycle                & Ring of Trees        \\
                                             & $|E| = 95, |V| = 96$ & $|E| = 96, |V| = 96$ & $|E| = 96, |V| = 96$ \\ \midrule
$\R^6$                                       & $0.1036$             & $0.1042$             & $0.1060$             \\
$\mathbb{H}^{-1, 6}$                         & $\mathbf{0.0454}$             & $0.2356$             & $0.0736$             \\
$\mathbb{S}^{1, 6}$                          & $0.1440$             & $\mathbf{0.0011}$             &         $0.1365$              \\
$\mathbb{H}^{-1, 3}\times \mathbb{S}^{1, 3}$ & $0.0624$             & $0.1337$             & $\mathbf{0.0686}$             \\ \bottomrule
\end{tabular}%
}
\label{tab:avg_distortion}
  \end{minipage}
  \vspace{-10pt}
\end{wraptable}

\textbf{Empirical Validation.} We empirically validate our claim that Euclidean space fails to capture complex structures faithfully and that non-Euclidean spaces are better suited for producing high-quality embeddings. Table~\ref{tab:avg_distortion} compares the average (point-wise) distortion of four geometric spaces (\( \mathbb{R}^6 \), \( \mathbb{H}^{-1, 6} \), \( \mathbb{S}^{1, 6} \), and \( \mathbb{H}^{-1, 3} \times \mathbb{S}^{1, 3} \)) in representing three canonical graphs (Tree, Cycle, and Ring of Trees) with 96 nodes, each corresponding to a different type of intrinsic token relationships (hierarchical, cyclical, and both). We intentionally dismissed invariance from kernel designs—both Euclidean and non-Euclidean, e.g. orthogonal invariance (see \cref{sec:3.3})—to isolate the representational capacity of the spaces. The most suitable geometry varies by graph type---Lorentzian space (\( \mathbb{H}^{-1, 6} \)) for trees, spherical space (\( \mathbb{S}^{1, 6} \)) for cycles, and mixed geometry (\( \mathbb{H}^{-1, 3} \times \mathbb{S}^{1, 3} \)) for rings of trees---emphasizing the importance of selecting an appropriate geometry to minimize distortion. As a fundamental desired property of foundation models is for them to generalize to a wide variety of data, non-Euclidean geometry enables significantly increased flexibility in the choice of geometry for embedding spaces, such as the use of a product of manifolds of varying curvatures (positive, zero, and negative) that encompasses the Euclidean space.

\begin{wrapfigure}{r}{0.5\textwidth}
\vspace{-12pt}
\centering
    \includegraphics[width=0.49\textwidth]{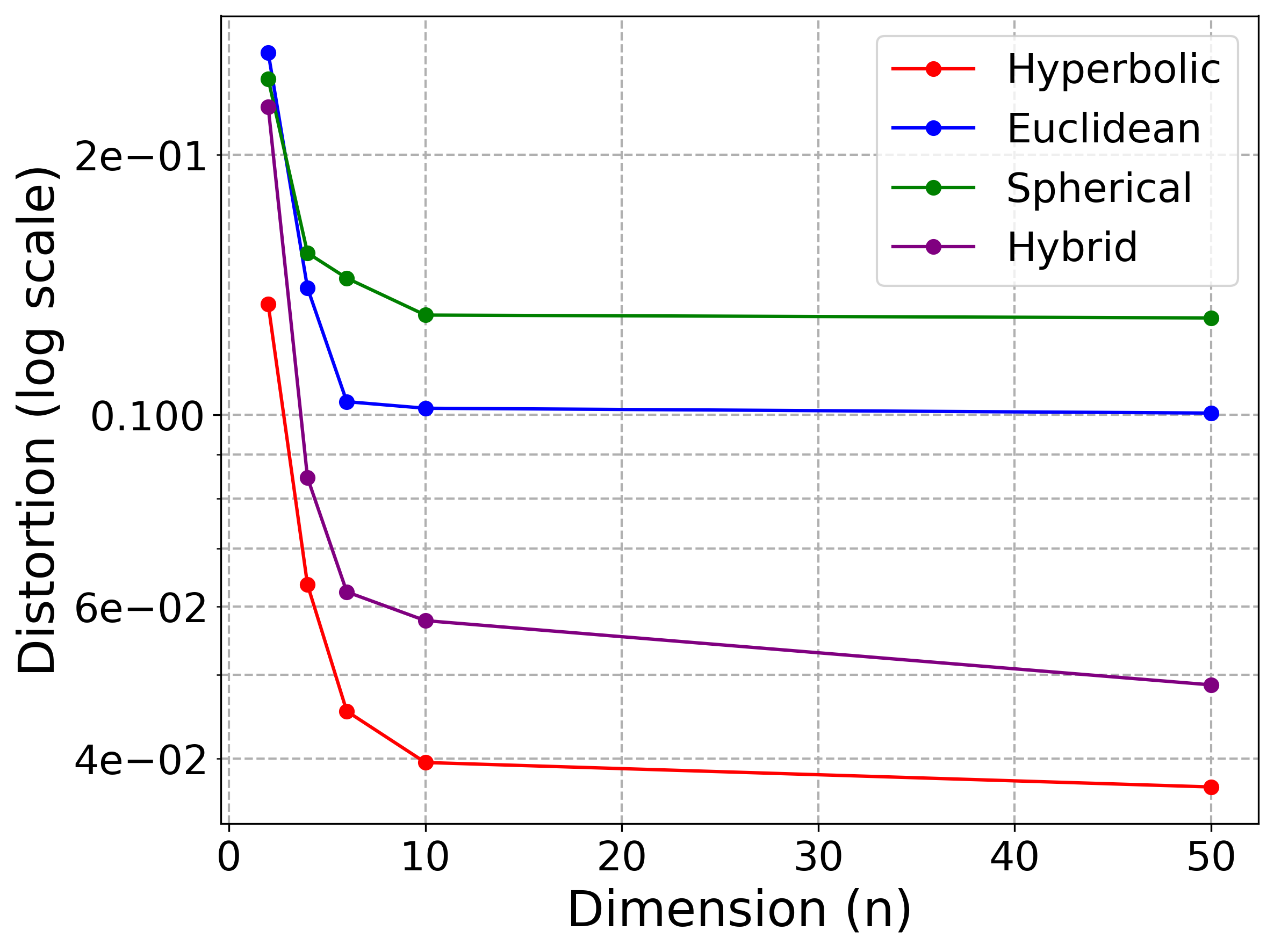}
    \vspace{-8pt}
    \caption{\small Distortion for embedding a Tree with 96 nodes for varying dimensionality (log scale). Non-Euclidean geometry achieves smaller distortion with significantly fewer dimensions and has better scaling.
}
    \label{figure:distortion_vs_dimension_tree}
    \vspace{-10pt}
\end{wrapfigure}

We also compute the distortion value against varying dimensionality. An example is shown in \cref{figure:distortion_vs_dimension_tree} for the case of a tree with 96 nodes, plotted on log-scale for visibility. The hybrid manifold is 
a product of hyperbolic and spherical spaces, each with half the dimension. The 4-dimensional hyperbolic space achieves a significantly smaller distortion than Euclidean embeddings with 50 dimensions. \textbf{This reflects Takeaway 1 in \cref{sec:euclidean_limitation}}: non-Euclidean geometry achieves superior performance with significantly fewer dimensions. Distortion also continues to decrease for hyperbolic and hybrid spaces but plateaus for Euclidean space, \textbf{reflecting Takeaway 2}, where Euclidean space has theoretical upper bounds for embedding trees but non-Euclidean geometry has the potential to continue the performance scaling law at high dimensionality. See \cref{additional_statistics} for additional plots of other graph types.

\textbf{Hierarchies in Token Embeddings.} Based on the above results validating Euclidean embeddings' limitations, we further show that these structures which Euclidean spaces struggles to embed is prevalent in foundation models. To explore the intrinsic structure within the representations of foundation models, we utilize \textbf{$\delta$-hyperbolicity}~\cite{gromov1987hyperbolic}, which quantifies the extent to which a metric space deviates globally from a tree metric (see \cref{appendix:delta_hyperbolicity}). Each token is treated as a point in a discrete metric space $X$, and a graph is constructed based on similarity scores between each pair. We evaluate the hyperbolicity of token embeddings in LLMs, where lower values suggest a tree-like structure. As shown in \cref{hyperbolicity}, the consistently low $\delta$-hyperbolicity values suggest \textit{hierarchical structures}  within each prompt across diverse datasets. 

We also analyze the global token embedding distribution in LLMs and pre-trained vision models using datasets included in the models' training corpus~\cite{touvron2023llama, dubey2024llama3, sun2024autoregressive}. \cref{fig:token_distribution} plots token input frequency distributions and their occurrences in the dataset on a log scale, revealing a \textit{scale-free} structure among the token embeddings. This scale-free organization suggests an underlying hierarchical structure~\cite{barabasi1999emergence}, where a small number of high-frequency tokens act as hubs within the semantic network. The figure also shows token norm distributions for learned embeddings, where the count for high-norm embeddings increases exponentially at the left tail, reinforcing the scale-free property. The non-Euclidean structures in token distribution are exhibited to some extent even in Euclidean models are most likely attributed to the models being optimized during training to maximize representational quality.
However, \textit{the scale-free properties are still not yet fully captured by the Euclidean foundational model, where the count of embeddings with large norms still decreases at the right tail.}  See \cref{additional_statistics} for more statistics. 

\textbf{Token Hierarchies to Model Performance.} While limited works have explored how embedding distortions connect to model performance and concrete connection has not been established due to a variety of challenges (see \cref{alternative views}), recent works have made preliminary efforts toward this goal. Recent efforts have shown that hyperbolic LLMs outperform Euclidean counterparts at multi-choice question answering tasks while exhibiting better semantic hierarchy modeling and better separation of words based on specificity, where more specific words are embedded further away from the origin~\cite{he2025helm}. These results suggest that
the failure of Euclidean geometry to model complex structure is a potential cause of degraded downstream performance.

\textbf{Additional Structures.} In addition to hierarchical structure, data may exhibit other structural characteristics, such as cycles and loops. 
Many real-world tasks, such as 3D shape analysis~\cite{esteves2017learning, esteves2019equaivariant}, medical imaging~\cite{bekkers2018roto, winkels20183d}, and physics-informed machine learning~\cite{liao2023equiformer, equiformer_v2, anderson2019cormorant, cohen2018spherical}, can benefit from encoding data geometry as inductive bias. Euclidean operations, such as convolutional layers, encode only translation invariance~\cite{esteves2020swscnn}, resulting in performance limitations for these tasks.


\subsection{The Necessity of Non-Euclidean Geometry for Foundation Models}\label{sec:3.3}
Here we further explore how non-Euclidean geometry could improve foundation model performance.

\textbf{(1) Addressing the limitations in capturing intrinsic token structures}.
Recent research shows that the attention mechanism plays a pivotal role in the expressive capacity of LLMs~\cite{aghajanyan2021intrinsic, song2023uncovering, vaswani2017attention, balestrierocharacterizing}. 

\begin{lemma}[\citet{balestrierocharacterizing}]
Let \( X \in \mathbb{R}^{T \times D_{(\ell)}} \) be the input to the \( \ell \)-th layer of an LLM, where \( T \) is sequence length and \( D_{(\ell)} \) is feature dimension. Attention head \( h \)'s output  at position \( i \) is in the convex hull of the first \( i \) rows of \( X V_{h,{(\ell)}} \):
$
\text{Head}_{h,{(\ell)}}(X)_i \in \text{Hull} \left\{ (V_{h,{(\ell)}})^\top x_j \mid j = 1, \dots, i \right\}.
$ with bounded effective dimension: $
\text{dim}_{\text{eff}} \leq \# \left\{ \text{Attn}_{h,{(\ell)}}(X)_{i,j} > 0 \mid j \in \{1, \dots, i\} \right\}.
$
Here, \( \text{Attn}_{h,{(\ell)}}(X) \) is the attention matrix for head \( h \) at layer \( \ell \):
$
\text{Attn}_{h,{(\ell)}}(X) = \text{softmax}_{\text{causal}}(X Q_{h,{(\ell)}} K_{h,{(\ell)}}^\top X^{\top}).
$

\end{lemma}
This lemma \textit{highlights that next-token prediction in LLMs is strongly influenced by relationships encoded in previous tokens.} As shown in Table~\ref{hyperbolicity}, tokens exhibits non-Euclidean characteristics. Consequently, the standard Euclidean attention mechanism does not faithfully capture hierarchical syntax, periodic dependencies, and other complex token relationships, as demonstrated in \cref{sec:euclidean_limitation}.
Utilizing non-Euclidean attention mechanisms instead could \textit{better capture previous token relationships by aligning with the intrinsic data structure, thus enhancing next-token prediction.} For example, hyperbolic geometry compresses distances exponentially, ensuring that distant but structurally related tokens (e.g., a root concept and its distant co-occurrences in a prompt) remain meaningfully close, enabling attention mechanisms to \textit{efficiently capture long-range dependencies and hierarchies.}

\textbf{(2) Alleviating distortion-dimension trade-offs.}
Recent studies examined how Euclidean-based LLMs encode hierarchies geometrically~\citep{park2024geometry, parklinear}, where a mapping function \( \lambda \) maps input text \( x \) to a vector \( \lambda(x) \in \mathbb{R}^d \), and an un-embedding layer assigns \( \gamma(y) \in \mathbb{R}^d \) to each token \( y \). The token probability distribution is given by
$
P(y \mid x) = \frac{\exp(\lambda(x)^\top \gamma(y))}{\sum_{y' \in \text{Vocab}} \exp(\lambda(x)^\top \gamma(y'))}.
$
To unify the different spaces, the embedding and unembedding spaces can be reformulated using transformations
$
g(y) = A(\gamma(y) - \bar{\gamma}_0), \quad \ell(x) = A^{-\top} \lambda(x),
$
where the Euclidean inner product serves as the causal inner product. This framework shows that Euclidean LLMs encode hierarchical concepts \textit{orthogonally}, where parent (e.g., animal) and child (e.g., bird, mammal) vectors are perpendicular. Yet, since 
 provides only d orthogonal dimensions, Euclidean spaces must rely on \textit{high dimensionality} to capture the expansive semantic hierarchies of language~\cite{gupta1999embedding}. \textit{Non-Euclidean spaces offer a more efficient alternative, preserving hierarchical relationships while significantly reducing dimensionality~\cite{nickel2017poincare, nickel2018learning}.}

\textbf{(3) Improved multi-modal heterogeneity modeling.} 
Data from different modalities vary significantly due to contextual factors, use cases, cultural differences, and different interpretations of the same information.
This complexity intensifies in multi-modal data, where each modality has distinct complex structures~\cite{li2022clmlf, wei2023tackling, liang2022mind, goel2022cyclip, jiang2023understanding}. For instance, latent modality gap and distinct modality structures exist in the latent space due to initialization and the contrastive learning process, impacting downstream tasks~\cite{liang2022mind}.
Different modalities also exist on separate manifolds~\cite{wangexploring}, making a unified Euclidean foundation model \textit{highly redundant in parameters and requiring varying degrees of pruning for different modalities.} Thus, Euclidean space struggles to capture multi-modal cross-domain relationships, as its flat structure lacks the flexibility needed for multi-faceted interactions in the data.

Non-Euclidean spaces exhibit much more geometric flexibility to enable multiple manifolds that encode different data distributions~\citep{wilson2014spherical, gao2022curvature, gao2021curvature}. For instance, hyperbolic geometry excels in vision-language foundation models by effectively capturing hierarchical relationships~\cite{desai2023hyperbolic, pal2024compositional, ramasinghe2024accept}, improving performance in tasks such as image-video-skeleton~\cite{li2024isolated} and video-audio applications~\cite{hong2023hyperbolic} while enhancing representation interpretability---higher-level hierarchical concepts lie closer to the origin with more specific concepts residing in more peripheral regions, enabling geodesic reasoning when navigating through concept hierarchies. 

\begin{figure*}
    \centering
\includegraphics[width=0.95\linewidth]{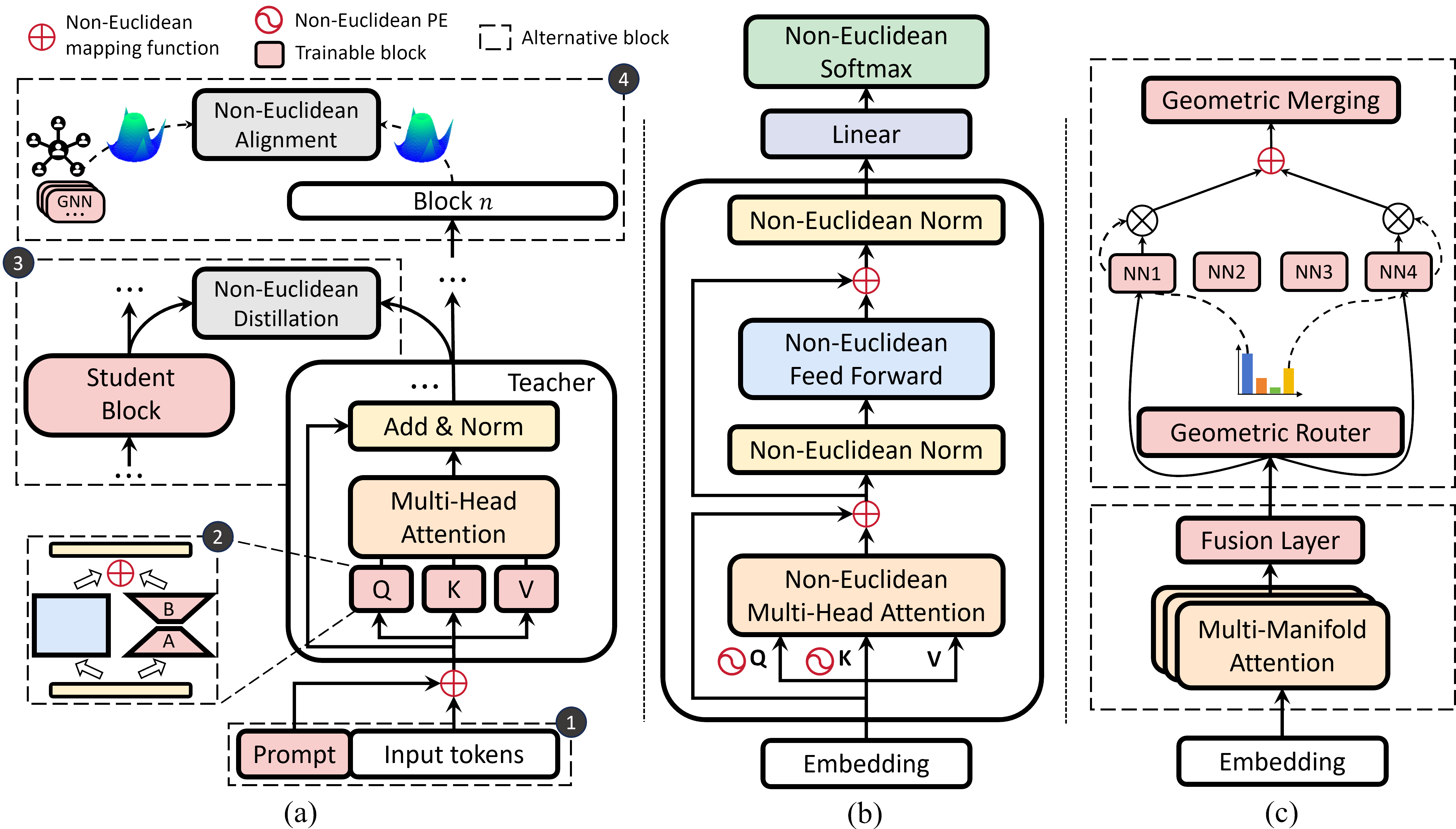}
   \caption{
   Roadmap for integrating non-Euclidean geometries into foundation models, includes (a) fine-tuning existing Euclidean foundation models, (b) pretraining from scratch, and (c) hybrid architectures. Four strategies are shown in (a), labeled with circled numbers 1-4, respectively: geometric prompt tuning, geometric low-rank adaptation, geometric knowledge distillation, and geometric transfer learning. All learnable components are highlighted in red in (a) and (c). 
   }
   \label{model_arch}
   \vspace{-10pt}
\end{figure*}

\section{Towards Non-Euclidean Foundation Models}

We propose a roadmap that explores three progressive approaches to incorporate non-Euclidean geometry in foundation models: fine-tuning existing Euclidean models, building non-Euclidean models from scratch, and developing a hybrid framework combining both for optimal performance. These methods could be evaluated with standard 

\subsection{Fine-tuning Existing Euclidean Foundation Models}

Off-the-shelf pre-trained Euclidean foundation models are strong starting points as they already encode rich information. An efficient strategy is to adapt them to non-Euclidean spaces, thereby retaining their original capabilities and enabling generalization to data with non-Euclidean structures. We propose four strategies, shown in Figure~\ref{model_arch}(a): (1) Geometric prompt tuning; (2) Geometric low-rank adaptation; (3) Geometric knowledge distillation; and (4) Geometric transfer learning.

\textbf{(1) Geometric Prompt Tuning.}
Prompt tuning offers a parameter-efficient alternative to full fine-tuning by introducing trainable, task-specific prompt tokens to the input, mitigating catastrophic forgetting while requiring fewer trainable parameters.~\cite{kirkpatrick2017overcoming, han2024parameter}. Geometric prompts can be optimized through non-Euclidean spaces to better align with the data geometry to and adapt to downstream tasks. For instance, trainable prompt and token embeddings could be introduced to better capture the topological relationships between prompts and text inputs.

\textbf{(2) Geometric Low-Rank Adaptation.}
Low-Rank Adaptation (LoRA) offers an efficient way to adjust the model parameter space for downstream tasks~\cite{hu2021lora}.
To equip the pre-trained model with non-Euclidean geometry through geometric low-rank adaptation, low-rank matrix multiplications could be performed directly on the manifold after projecting the input into non-Euclidean spaces, which better models the underlying geometric structure of the data~\cite{yang2024hyperbolic}. 

\textbf{(3) Geometric Knowledge Distillation.}
Distilling knowledge into non-Euclidean spaces refers to transferring knowledge from a large, complex teacher model to a smaller, more efficient student model by utilizing manifold properties to teach the student to better inherit the teacher model's geometric structure. An example is minimizing the gap between each layer's output of both models, especially in high-dimensional spaces~\cite{yang2022geometric,hao2022learning} and resource-limited applications.

\textbf{(4) Geometric Transfer Learning.} Geometric transfer learning aims to help foundation models learn across domains with aligned geometries, ensuring a much more effective and consistent knowledge transfer. Geometry alignment objectives can be designed to supervise the transfer of geometric knowledge, such as hyperbolic contrastive learning for recommendation~\cite{10.1145/3627673.3679572, ma2024harec}, preserving the intrinsic structure of the target domain while retaining geometry-agnostic prior knowledge. 

\subsection{Pretraining from Scratch}

Pretraining non-Euclidean foundation models requires addressing unique challenges. We outline key components for adapting models to complex curvature-aware structures; see also Figure~\ref{model_arch}(b). A detailed mathematical formulation is presented in Table~\ref{tab:formula} in the Appendix.

\textbf{Curvature Estimation.} 
A manifold's curvature determines its intrinsic geometric properties, such as distance metrics and learning dynamics.
Curvature estimation methods vary based on data types. For \textit{graph data} (e.g., networks, proteins), curvature can be derived from topological properties, such as Ollivier-Ricci curvature or Gromov hyperbolicity~\cite{ollivier2009ricci,jost2014ollivier,yang2024hyperbolic,trillos2023continuum}. For \textit{non-graph data} (e.g., texts, images), curvature can be estimated from learned embeddings~\cite{khrulkov2020hyperbolic, atigh2022hyperbolic} or techniques like Isomap~\cite{tenenbaum2000isomap} and UMAP~\cite{mcinnes2018umap}. One could also design learnable curvature within training pipelines using second-order statistics~\cite{gao2022curvature}, reinforcement learning~\cite{fu2021ace}, and self-supervised learning~\cite{sun2023self,sun2022self}.

\textbf{Non-Euclidean Attention Mechanism.} In non-Euclidean spaces, attention scores can be defined based on negative manifold distance $-d_\mathcal{M}(x,y)$ between queries and keys instead of dot products~\cite{gulcehre2019hyperbolicAT, HNN++, chen2021fully, cho2023curve}, with closer node pairs receiving higher attention weights. To aggregate attention, unified manifold centroids or tangent space operations can be used~\cite{gu2019mixedcurvature, cho2023curve}. Linear attention mechanisms~\cite{yang2024hypformer} can be employed to improve computational efficiency by approximating traditional attention through unified tangent space operations.

\textbf{Other Important Modules.}
Traditional Euclidean positional encodings~\cite{vaswani2017attention, su2021roformer} do not preserve the manifold structure in non-Euclidean spaces. Several approaches for non-Euclidean positional encoding~\cite{chen2021fully, yang2024hypformer, fein-ashley2024hvt} were proposed to represent token positions while maintaining geometric integrity. \emph{Residual connections} should be formulated using isometric operations~\cite{he2025lresnet, van2023poincar, Bdeir2024fully} to preserve geometric information across layers. Layer and batch normalization must also be adapted to account for curvature~\cite{yang2024hypformer, Bdeir2024fully, van2023poincar}. Loss function must also satisfy geometric constraints, such as computing the probability distribution over tokens based on the manifold distance instead.

\subsection{Hybrid Architectures}
Hybrid architectures take a step further by merging both
Euclidean and non-Euclidean foundation model architectures to provide a more universal inductive bias. We illustrate two promising strategies, also depicted in Figure~\ref{model_arch}(c).

\textbf{Dynamic Geometry Adaptation.}
An intuitive way for hybrid modeling is to design an efficient and geometry-aware mechanism that shifts dynamically between manifolds. 
Unified product manifold frameworks~\cite{10.1609/aaai.v38i8.28754} could enable layers to integrate diverse learnable curvature values that adapt to fine-grained geometric structures. Mixture of Experts (MoE)~\cite{zhou2022mixture} 
provides a natural framework for hybrid paradigms to use a geometry-aware sparse routing network by selecting the most appropriate geometry considering input structure~\cite{guo2024graphmoremitigatingtopologicalheterogeneity}, addressing issues of distortion and heterogeneity.

\textbf{Multi-Manifold Attention.}
Multi-manifold attention could lead to more versatile underlying dependencies
~\cite{konstantinidis2023multi,kalitsios2024dynamic}, where the input is embedded into a collection of manifolds (including Euclidean) to represent differences in geometric structure across the dataset. These geometric attention maps are then fused to produce a highly discriminative map for improved attention guidance.

\section{Alternative Views, Implementation Challenges, and Potential Solutions}\label{alternative views}
\textbf{Engineering Overhead.} 
While non-Euclidean geometries have clear theoretical benefits, their operations can introduce significant engineering overhead that may offset efficiency gains. However, as mentioned earlier in the paper, non-Euclidean models require fewer dimensions to embed complex structures, as seen in our discussion in \cref{sec:euclidean_limitation} and \ref{sec:geometry_stats}. This enables the potential for non-Euclidean models to match the performance of Euclidean models with fewer parameters to offset the computational overhead while offering additional benefits, such as the potential to continue the scaling law relationship between parameters and model performance. 

Nevertheless, designing efficient non-Euclidean operations is therefore a critical direction for future work. In particular, tangent-space-enabled methods~\cite{HNN, hgcn2019, van2023poincar} incur significant computation overhead due to multiple mappings to and from the tangent bundle. In comparison, methods that operate directly on the manifold~\cite{chen2021fully, yang2024hypformer, he2025lresnet}, while still computationally more expensive than Euclidean methods, typically have similar computational complexity as their Euclidean counterparts. Thus, they could be promising for managing the computational efficiency of non-Euclidean foundation models. Additionally, it is essential to develop libraries, such as~\cite{he2025hypercore} that optimize these computations, with efficient implementations of tensor operations that encode the underlying geometry, geometric optimization techniques tailored to non-Euclidean spaces, and multi-node training architectures such as non-Euclidean flash attention to account for architectural differences as a result.

\textbf{Learning Inductive Bias with Euclidean Models.} Another family of views is to let Euclidean foundation models learn geometric properties rather than hard-coding inductive biases. One argument holds that with increasing hardware capacity, scaling Euclidean models to higher dimensions could reduce distortion and capture geometry, as seen in CNNs and ViTs~\cite{dosovitskiy2020image}. A complementary argument is that improved data quality and quantity might allow Euclidean models to learn sufficient geometric structure. However, as shown in \cref{dist_theorem} and \ref{sphere_dist}, many desirable geometric biases cannot be faithfully learned by Euclidean foundation models. Previous works have also empirically shown that non-Euclidean models outperform Euclidean models even with scaled parameter counts, such as for equivariant and non-equivariant models~\cite{brehmer2024does}.  Additionally, in many domains, such as molecular structures or rare languages, data scarcity results in brute-force scaling being ineffective. Non-Euclidean geometries, on the other hand, can capture important relationships even in lower-dimensional settings~\cite{sarkar2011low}, making them efficient in data requirements, offering better performance scalability w.r.t. model size, and are more reliable for domains with limited high-quality data. 

\textbf{Distortion-Performance Connection.} Additional views include gaps in analyzing the link between embedding quality and downstream foundation model performance. Prior work suggests that manifolds better capturing data structure can improve graph tasks and word embeddings~\cite{gu2019mixedcurvature}. Better semantic hierarchy modeling by incorporating non-Euclidean geometry could also lead to improved performance in LLMs and VLMs~\cite{he2025helm, desai2023hyperbolic}. However, to our knowledge, no conclusive studies connect distortion directly to downstream performance—a direction we advocate in this paper. This is challenging as it could require prior knowledge of the ground-truth data geometry, compute resources to train multiple foundation models, and isolating the effects of distortion. Future works in this aspect would provide valuable insights to support better development of non-Euclidean methods for foundation models. Additional analysis that could aid in this direction include analyzing embedding distribution in non-Euclidean models.

\section{Conclusion}
Foundation models benefit from embracing non-Euclidean geometry to resolve their inherent mismatch with the non-Euclidean nature of real-world data. Non-Euclidean geometries reduce distortion for embedding complex structures and relationships while enabling efficient representations, which is critical for trillion-parameter scaling. Aligning architectures with data geometry could mitigate hallucinations, boost efficiency, and unlock heterogeneous scaling. We encourage the community to consider three directions: unified curvature-adaptive foundation models, geometry-aware benchmarks, and studying manifold-emergent capability links. Embracing this paradigm will catalyze AI systems that better reflect the rich geometries of human knowledge and physical reality.

\section*{Acknowledgments}
This work was supported in part by the National Science Foundation (NSF) IIS Div Of Information \& Intelligent Systems 2403317. We also acknowledge support in part from the Silicon Valley Community Foundation, an Amazon research award, the Yale AI Engineering Research Grant from Yale Office of the Provost, and an LEAP-U Sponsored Research from Samsung Research America. Moreover, this research has greatly benefited from the discussions and research talks held at the IMS-NTU Joint Workshop on Applied Geometry for Data Sciences. 

\bibliography{reference}
\bibliographystyle{plainnat}
\newpage
\appendix

\section{Comprehensive Background and Related Works}\label{extended_related_work}

\subsection{Riemannian Geometry and Non-Euclidean Foundations}

\textbf{Riemannian Manifolds.} A smooth $n$-dimensional manifold $\mathcal{M}$ is a topological space in which each point $x \in \mathcal{M}$ has a neighborhood $U_x \subseteq \mathcal{M}$ that is locally Euclidean, meaning that there exists a homeomorphism between $U_x$ and an open subset of $\mathbb{R}^n$.

\textbf{Tangent Space.} Each point $x \in \mathcal{M}$ is associated with a \textit{tangent space} $T_x \mathcal{M}$, which is an $n$-dimensional vector space serving as a first-order local approximation of $\mathcal{M}$ at $x$. This space encapsulates the possible directions in which one can move away from $x$ on the manifold.

\textbf{Riemannian Metric.} A \textit{Riemannian metric} $\mathfrak{g}$ on $\mathcal{M}$ is a collection of positive-definite bilinear forms $\mathfrak{g}_x(\cdot, \cdot): T_x \mathcal{M} \times T_x \mathcal{M} \to \mathbb{R}$, smoothly varying with $x$. The metric $\mathfrak{g}_x$ induces the \textit{sectional curvature} at each point, which measures the extent to which the manifold deviates from flatness at $x$. A \textit{Riemannian manifold} is then defined as the pair $(\mathcal{M}, \mathfrak{g})$. For instance, $\mathbb{R}^n$ with the usual Euclidean inner product is a Riemannian manifold with zero curvature. The metric $\mathfrak{g}_x$ generalizes the notion of inner products, with the norm of a vector $p \in T_x \mathcal{M}$ given by $\|p\|_{\mathfrak{g}} = \sqrt{\mathfrak{g}_x(p, p)}$. The choice of the Riemannian metric also induces a global distance function $d(\cdot, \cdot)$ on $\mathcal{M}$.

\textbf{Geodesic.} A \textit{geodesic} between two points $x$ and $y$ is a smooth curve that locally minimizes the distance between these points. In particular, the shortest path between $x$ and $y$ is a geodesic.

\textbf{Exponential Map.} Under certain conditions, one can define the \textit{exponential map} $\exp_x: T_x \mathcal{M} \to \mathcal{M}$, which lifts points from the tangent space $T_x \mathcal{M}$ to the manifold $\mathcal{M}$, by associating a vector in $T_x \mathcal{M}$ to a point on $\mathcal{M}$ along a geodesic.

\textbf{Logarithmic Map.} The \textit{logarithmic map} $\log_x: \mathcal{M} \to T_x \mathcal{M}$ is the inverse of the exponential map, provided certain assumptions on $\mathcal{M}$ hold. 

\textbf{Geodesics and Geodesic Operations.} The Riemannian metric $\mathfrak{g}_x$ can be viewed as a generalization of the inner product, where the norm of a vector $p \in T_x \mathcal{M}$ is defined by $\|p\|_{\mathfrak{g}} = \sqrt{\mathfrak{g}_x(p, p)}$. The choice of $\mathfrak{g}$ induces a global distance function $d(\cdot, \cdot)$ on $\mathcal{M}$, where geodesics are the locally distance-minimizing curves. The length of a geodesic between two points determines the geodesic distance. The exponential map $\exp_x$ maps a vector $v \in T_x \mathcal{M}$ to a point on $\mathcal{M}$ along the geodesic starting at $x$. The logarithmic map $\log_x$ is the inverse of this process. Additionally, the parallel transport map $\mathrm{PT}_x(v, w)$ transports vectors along geodesics, providing a canonical way to move vectors in a manner consistent with the underlying geometric structure. It canonically transports a vector $w$ along a geodesic emanating from $x$ with initial velocity $v$ and zero acceleration. This generalizes the classical notion of translation in Euclidean space.

\textbf{Hyperbolic Spaces.} Hyperbolic spaces are Riemannian manifolds with constant negative curvature, i.e., with curvature $-K < 0$. Common models for hyperbolic space include the \textit{Poincaré ball model} $\mathbb{P}^{K, n}$ and the \textit{Lorentz hyperboloid} $\mathbb{L}^{K, n}$, which have been extensively studied in the context of deep learning~\cite{nickel2017poincare, HNN}. For points $\mathbf x,\mathbf y\in\mathbb{L}^{K,n}$, their inner product $\langle\mathbf x,\mathbf y\rangle_\mathcal{L}$ is given by $\langle\mathbf x,\mathbf y\rangle_{\mathcal{L}} = -x_ty_t + \mathbf x_s^T \mathbf y_s = \mathbf x^T\mathfrak{g}_n^K\mathbf y$
with $|\|\mathbf x\||_\mathcal{L}\coloneq\sqrt{|\langle \mathbf x, \mathbf x\rangle_\mathcal{L}|}$ being the Lorentzian norm. Formally, $\mathcal{L}^n$ is the point set $\mathcal{L}^n = \{\mathbf x\in\mathbb R^{n+1}: \langle\mathbf x,\mathbf x\rangle_\mathcal{L} = 1/K, x_t>0\}.$ $\mathbb{P}^{n, K}$ is the $n$-dimensional sphere $S^n$ with radius $1/\sqrt{K}$, with the Riemannian metric $g_x^{\mathbb{P}} = \lambda_x^2 g^E$, where $ \lambda_x := \frac{2}{1 - c\|x\|^2}$ and $\mathfrak{g}^E$ is the Euclidean metric. Other models, such as the Klein model, also exist. These models are \textit{isometric}, meaning that there is a smooth correspondence between points in different models that preserves distances, angles, and geodesics. This property allows for the selection of the most suitable model for a given application.

\textbf{Spherical Spaces.} Spherical spaces are Riemannian manifolds with constant positive curvature, i.e., with curvature $K > 0$. An $n$-dimensional spherical space $\mathbb{S}^{K, n}$ is an $n$-dimensional sphere of radius $K^{-\frac{1}{2}}$, equipped with the Riemannian metric induced by the Euclidean metric on $\mathbb{R}^{n+1}$.

\textbf{Mixed Curvature Spaces.} A \textit{mixed curvature space} $\mathcal{M}$ is defined as a product manifold consisting of Euclidean, spherical, and hyperbolic spaces. The Riemannian metric and geodesic operations for such a manifold are defined component-wise, enabling effective computational implementation for downstream tasks.

\textbf{Generalized Riemannian Manifolds.} Generalizations of Riemannian manifolds can be obtained by relaxing some of the assumptions in their classical definition. One notable generalization is the \textit{pseudo-Riemannian manifold}, in which the metric $\mathfrak{g}$ is an indefinite bilinear form, allowing for both positive and negative signs. This generalization is useful in contexts such as relativistic physics, where spacetime is modeled as a pseudo-Riemannian manifold.

\textbf{Geometry of Embedding Spaces.} Geometric assumptions are not only the result of neural network operations but are also inherently present when one chooses the latent space, whose underlying geometric properties determines not only how distances between representations are defined, but also what kind of operations are applicable in it. Consider a simple MLP example. In this model, each layer applies a linear transformation (e.g., $Ax + b$) followed by a non-linearity. Just by applying these transformations, one is implicitly assuming that the MLP's representations lie in a geometrically Euclidean space that is flat and globally uniform, so that operations such as addition and scaling make sense everywhere in the same way. Simply put, one is implicitly assuming that the geometry of the space does not depend on location, and vectors can be freely added or transformed. In contrast, Riemannian manifolds are defined by a smooth manifold together with a Riemannian metric tensor. The smooth manifold defines the topological structure, which is not captured by the metric and is fundamentally not the same as just the set of $d$-dimensional real valued vectors (for example, for $d$-dimensional Lorentz hyperbolic space, $\mathcal{L} = \{x\in\R^{d+1}: \langle x,x\rangle_\mathcal{L}=1/K\}$), even though in reality one expresses the coordinates with real numbers. Thus, Riemannian manifolds have curvatures and do not, for example, support global linearity. This means that operations such as addition or scaling, which we take for granted in the standard MLP, cannot be directly applied on Riemannian manifolds. Instead, linear transformations must become location-dependent to account for curvature. Hence designing non-Euclidean neural networks typically requires redesigning the associated operations to conform to the underlying geometry. 

\subsection{Non-Applicability of the Nash Embedding Theorem}\label{nash_embedding}

The Nash Embedding Theorem roughly states that any Riemannian manifold of dimension $n$ admits an isometric embedding into $\R^{2n+1}$~\cite{nash1954embedding}. While it may appear as if this allows for Euclidean embeddings of complex structures with no distortion and only twice the dimension, this is in fact a confusion in vocabulary between the notion of isometric embeddings between those of \textit{Riemannian manifolds} and those of \textit{metric spaces}. 

\begin{definition}
    Let $(\mathcal{M}, \mathfrak{g}), (\mathcal{M}', \mathfrak{g}')$ be Riemannian manifolds. An \textit{isometric embedding of Riemannian manifolds} is a smooth map $f: \mathcal{M}\to\mathcal{M'}$ such that $\mathfrak{g} = f^*\mathfrak{g'}$. Let $(X, d_X), (Y, d_Y)$ be metrics spaces. An \textit{isometric embedding of metric spaces} is a map $f:X\to Y$ such that $d_X(a, b) = d_Y(f(a), f(b))$ for all $a,b\in X$.
\end{definition}

Hence in the former, which is also the isometric embedding afforded by the Nash Embedding Theorem, the map $f$ preserves the Riemannian metric, i.e. the inner product on the tangent bundle. As a result, the isometry is \textit{locally distance preserving}, in the sense that \textit{length of any path} is preserved under $f$. However, given points $x,y$ connected by a shortest path $\gamma$, the straight line path connecting $f(x), f(y)$ in the co-domain is not necessarily $f(\gamma)$ (note that $f$ need not to be surjective). As a result, measuring the distortion of embeddings is concerned with whether $f$ is \textit{globally distance preserving}, or whether the shortest distance between $f(x)$ and $f(y)$ is the length of $f(\gamma)$, which is defined by isometric embeddings between metric spaces. Note that \textbf{an isometric embedding of Riemannian manifolds is in general not an isometric embedding of metric spaces}. For instance, given the sphere $S^1$, its usual Riemannian metric is inherited from the Riemannian metric for $\R^2$, i.e. the usual inner product. The identity map is then an isometric embedding $S^1\hookrightarrow \R^2$ as Riemannian manifolds. However, the distance between points on the sphere does not coincide with the Euclidean distance of their image. As an example, antipodal points have distance $\pi$ in $S^1$ but distance $1$ in $\R^2$.   

In the context of foundational models, we are concerned with globally distance preserving embeddings, as computing the distance between any pairs of token embeddings is crucial for model training. As a result, the Nash Embedding Theorem is not applicable since global distortion could still arise from isometries between Riemannian manifolds. For this reason, by "isometry", we refer to those between metric spaces unless otherwise specified, which captures the notion of distortion critical for foundational model embeddings. 

\begin{definition}
    Let $(X, d_X), (Y, d_Y)$ be metric spaces equipped with the respective distance metrics and $f:X\to Y$ be a map. The \textit{bi-Lipschitz distortion} of $f$ is $\mathrm{dist}(f) = \|f\|_\mathrm{Lip}\|f^{-1}\|_\mathrm{Lip}$, where $\|f\|_{\mathrm{Lip}}$ is the (possibly infinite) Lipschitz-constant of $f$. For a pair of points $(a, b)\in X^2$, the \textit{point-wise distortion} is given by $\frac{|d_X(a,b) - d_Y(f(a)-f(b))|}{d_X(a,b)}$.
\end{definition}
Both notions of distortion measure the deviation of $f$ from an \textit{isometry between metric spaces}. Note that the minimum distortion in the case of bi-Lipschitz distortion is $1$. 

\subsection{Markov Convexity}\label{sec:markov}
In this section we provide the relevant background on notion of Markov convexity. Let $(X, d_X)$ be a metric space. Then the \textit{Markov $p$--convexity constant} $\Pi$ (for a fixed positive integer $p$) of the metric space $X$ is a universal constant (or $\infty$) define as follows:
\begin{definition}
   $\Pi$ is the smallest constant s.t. for any Markov chain on $(X_t)_{t\geq 0}$ on a state space $\Omega$, and every map $f:\Omega\to X$, and for any $m\in \mathbb{N}$, we have \[
   \sum_{n=0}^{\infty} \frac{1}{2^{np}}\sum_{t\in\mathbb{Z}}\mathbb{E}\!\Bigl[d\bigl(f(X_t),f(X_{t+2^n})\bigr)^p\Bigr]\;\;\le\;\;\Pi^p\sum_{t \in\mathbb{Z}}\mathbb{E}\!\Bigl[d\bigl(f(X_t), f(X_{t+1})\bigr)^p\Bigr]
   \]
\end{definition}
  Roughly speaking, when $\Pi<\infty$, the $p$-th moment of one-step increments dominates the $p$-th moments of exponential length steps. Intuitively, measures how tightly local behaviors in $X$ control and estimate global behaviors on the space, with lower values $\Pi$ showing tighter control. 

\subsection{Non-Euclidean Structure in the Real World}
\textbf{Non-Euclidean Structures in Natural Language Processing}. 
Language exhibits inherently hierarchical structures - from concept taxonomies to entailment relationships - that challenge traditional Euclidean representations. These hierarchical relationships between linguistic units naturally manifest on non-Euclidean manifolds, particularly in hyperbolic space, which has emerged as a powerful framework for natural language processing~\cite{dhingra2018embedding, leimeister2018skip}. 
Foundational work has demonstrated that hyperbolic embeddings can effectively capture word-level semantics~\cite{tifrea2019poincare} and concept hierarchies~\cite{le2019inferring}, leveraging the exponential volume growth of hyperbolic space to model tree-like linguistic structures. The success of hyperbolic representations has sparked various advanced applications: from question answering systems~\cite{tay2018hyperbolic}, privacy-preserving text representations~\cite{feyisetan2019leveraging}, to multi-document summarization that captures document-level discourse structure~\cite{song2023hisum}. Recent work has further extended these approaches to cross-lingual settings~\cite{saxena2022cross} and contextual language models~\cite{chen2021probing}, demonstrating the broad utility of non-Euclidean geometries in modern natural language processing.

\textbf{Non-Euclidean Structures in Computer Vision}.
Similar to NLP, many computer vision tasks involve data that naturally resides in intricate manifolds that are challenging to model using conventional Euclidean space~\citep{mettes2024hyperbolic}. For instance, visual entities often form inherent hierarchical relationships among object classes, between scenes and their constituent categories~\citep{ge2023hyperbolic, pal2024compositional}, or scenes at varying levels of granularity~\citep{khrulkov2020hyperbolic}. In these scenarios, hyperbolic geometry provides a compelling alternative to the Euclidean representations in representing the exponential growth of hierarchical structures with minimal distortion~\citep{sala2018representation}. Its advantages have been demonstrated across a wide range of applications, including image segmentation~\cite{atigh2022hyperbolic}, action classification~\citep{chen2022hmanet} video prediction~\cite{suris2021learning}, deformable 3D surfaces~\cite{masci2015geodesic}. In parallel, hyperspherical learning has become integral to modern contrastive learning with cosine similarity, underpining tasks ranging from self-supervised learning~\cite{durrant2022hyperspherically} to long-tailed classification~\cite{kasarla2022maximum} and few-shot learning~\cite{trosten2023hubs}.

\textbf{Non-Euclidean Structures in Complex Networks}.
Networks, whether they represent social interactions, user purchasing preferences, or transportation systems, often exhibit complex, non-Euclidean relationships that traditional Euclidean models fail to capture effectively. Social networks, for example, are best described by graph structures where nodes (individuals) are connected by edges (relationships) that can be directional, weighted, or even exhibit hierarchical properties. These networks typically involve intricate dependencies and nonlinear relationships, requiring geometric frameworks beyond Euclidean space to model effectively. 

\textbf{Non-Euclidean Structures in Natural Sciences}.
In natural science, many systems exhibit intricate structures that Euclidean space struggles to capture effectively. In biology, non-Euclidean geometries are integral to analyzing and modeling complex organic structures, such as protein folding~\cite{villegas2021foldhsphere}, single-cell RNA-seq data~\cite{klimovskaia2020poincare,ding2021deep,bhasker2024contrastive}, and phylogenetic trees~\citep{matsumoto2021novel}, where hyperbolic and spherical geometries are commonly observed. In neuroscience, hyperbolic geometry is shown to be more effective than Euclidean counterpart in modeling the brain's cortical folding~\citep{urdapilleta2015can}, brain surface~\cite{shi2013hyperbolic}, and hippocampal spatial representations~\citep{zhang2023hippocampal}, aiding in the study of spatial organization and connectivity.

\subsection{Deep Learning with Non-Euclidean Geometries}
Recent years have witnessed an increasing interest in extending deep learning techniques to Riemannian manifolds. Here we discuss in further detail the advances for designing neural networks and Transformers in non-Euclidean geometries, as well as optimization on manifolds.

\textbf{Geodesic Neural Networks.} Geodesic neural networks leverage geodesic, particularly geodesic distances, to perform neural operations that preserve geometric structure on manifold-structured data~\cite{bronstein2017geometric}. Several works have developed geodesic convolutional layers by applying filters to local patches in geodesic polar coordinates~\cite{masci2015geodesic}, learning directionally sensitive filters along principal curvature directions~\cite{boscaini2016learning}, or learnable kernel functions that operate on local coordinate systems~\cite{monti2017geometric}. More recent works such as GDGNN~\cite{kong2022geodesic} have integrated geodesic operations with graph representations.

\textbf{Hyperbolic Neural Networks.}
Hyperbolic neural networks exploit the geometry of hyperbolic space to learn embeddings that reflect hierarchical relationships more effectively than their Euclidean counterparts~\cite{nickel2017poincare}. HNN~\cite{HNN} and HNN++\cite{HNN++} developed many basic operations, such as hyperbolic linear and convolutional layers, and multinomial logistic regression (MLR). HGCN~\cite{hgcn2019} and HGNN~\cite{liu2019HGNN} were then among the first to develop hyperbolic graph neural networks (GNNs). 
More recently, HyboNet~\cite{chen2021fully} proposed a framework of hyperbolic neural networks that does not depend on the Euclidean tangent spaces; Poincar\'e ResNet~\cite{van2023poincar} and HCNN~\cite{Bdeir2024fully} developed components for hyperbolic vision models; LResNet~\cite{he2025lresnet} proposed an efficient and stable residual connection method. 

\textbf{Spherical Neural Networks.}
Spherical neural networks are designed for data that naturally reside on spheres or benefit from spherical symmetry. Spherical CNNs~\cite{cohen2018spherical, esteves2017learning} extended convolutions and pooling to preserve rotational symmetries. SphereNet~\cite{coors2018spherenet} introduced a framework for learning spherical image representations by encoding distortion invariance into convolutional filters. DeepSphere~\cite{deepsphere2020deepsphere} proposed a graph-based approach. SWSCNN~\cite{esteves2020swscnn} later proposed a fully spherical CNN that allows for anisotropic filters.

\textbf{Mix-curvature Neural Networks.}
Mix-curvature neural networks uses product spaces of the aforementioned manifolds to better model data that have local neighborhoods exhibiting different geometric properties. ~\cite{gu2019mixedcurvature} developed the first learning framework on product spaces, introducing fundamental techniques such as mean and loss functions for embedding optimization. $\kappa$-GCN~\cite{bachmann2020constant} then extended learning on product spaces to GCNs, introducing a unified and differentiable Gyrovector spaces framework to constant curvature spaces beyond hyperbolic manifolds.

\textbf{Non-Euclidean Transformers.}
Significant advancements have been made toward Transformers in non-Euclidean spaces in recent studies. Within hyperbolic learning, several works have proposed hyperbolic self-attention mechanisms~\cite{gulcehre2019hyperbolicAT, chen2021fully, HNN++} and hyperbolic linear attentions~\cite{yang2024hypformer}, enabling constructions of hyperbolic Transformers. Hyperbolic fine-tuning methods have also been developed for LLMs~\cite{yang2024hyplora}. Recent works have also proposed hyperbolic vision Transformers~\cite{fein-ashley2024hvt}. Attention mechanisms have been developed for spherical spaces as well~\cite{lai2023spherical}. Further, Transformers have been developed for mixed curvature manifolds as well~\cite{cho2023curve}.

\textbf{Optimization on manifolds.}
Learning on manifolds often times require optimizing parameters with manifold constraints. Many classical convex optimization algorithms have been extended to the manifold-valued setting~\citep{udriste1994convex,bacak2014convex,zhang2016first,weber2022riemannian}. Stochastic optimization on manifolds has been studied extensivley~\citep{bonnabel2013stochastic,zhang2016riemannian,becigneul2018riemannian,weber2021projection}, which includes extensions of algorithms such as SGD and Adam, which are suitable for training models on geometric domains.


\section{Additional Statistics and Dataset Details}
\label{additional_statistics}
In this section we give details regarding the datasets we used, as well as the show more statistic results for more LLMs. We also show the distortion v.s. dimensionality plot for all graph here.

\subsection{Distortion v.s. Dimensionality}
\begin{figure}
    \centering
    \begin{subfigure}[t]{0.40\textwidth}
        \centering
        \includegraphics[width=\linewidth]{Figures/distortion_scale_tree.png}
        \caption{Tree}
        \label{fig:t}
    \end{subfigure}
    \begin{subfigure}[t]{0.40\textwidth}
        \centering
        \includegraphics[width=\linewidth]{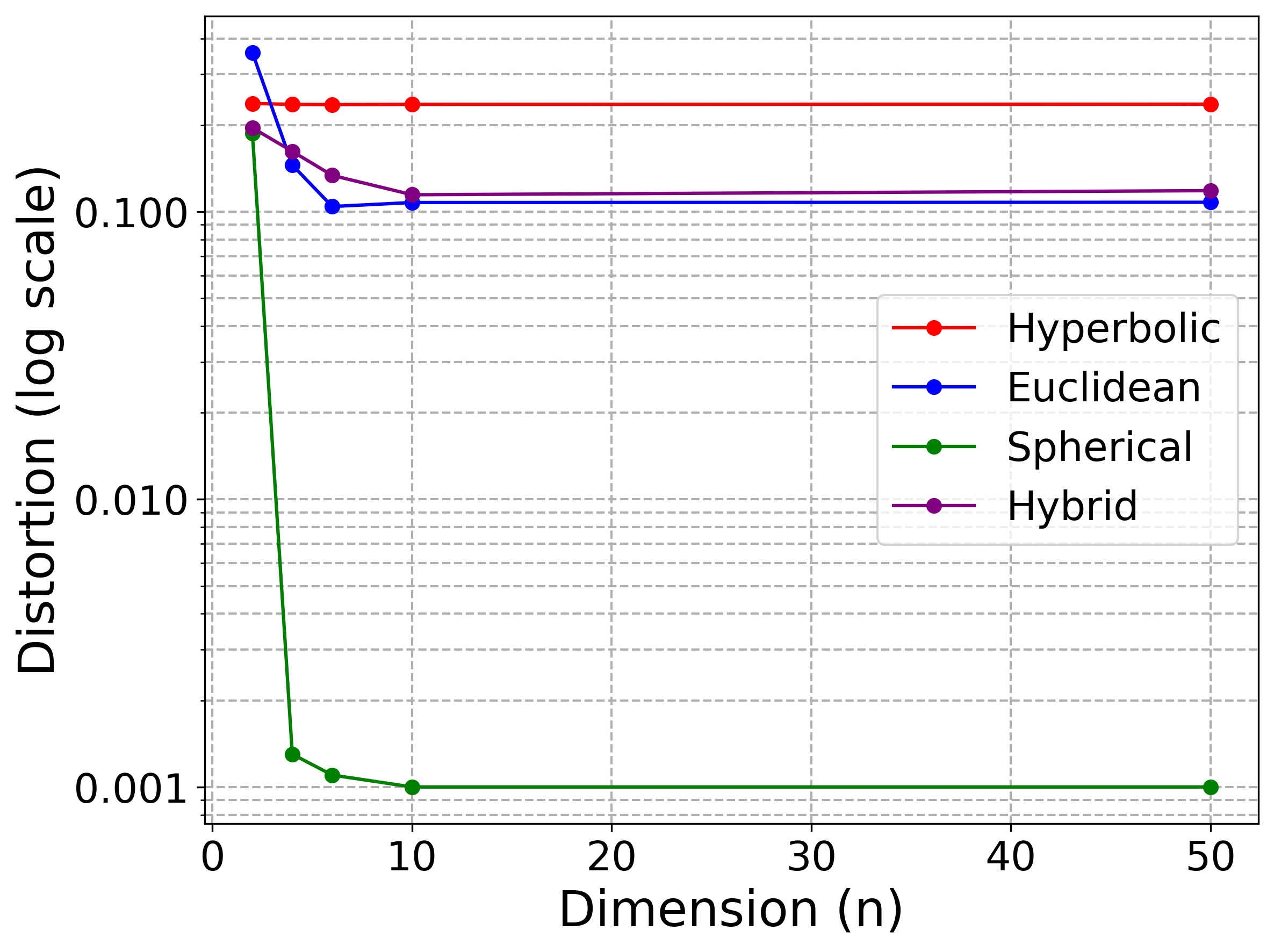}
        \caption{Cycle}
        \label{fig:c}
    \end{subfigure}
    \begin{subfigure}[t]{0.40\textwidth}
        \centering
        \includegraphics[width=\linewidth]{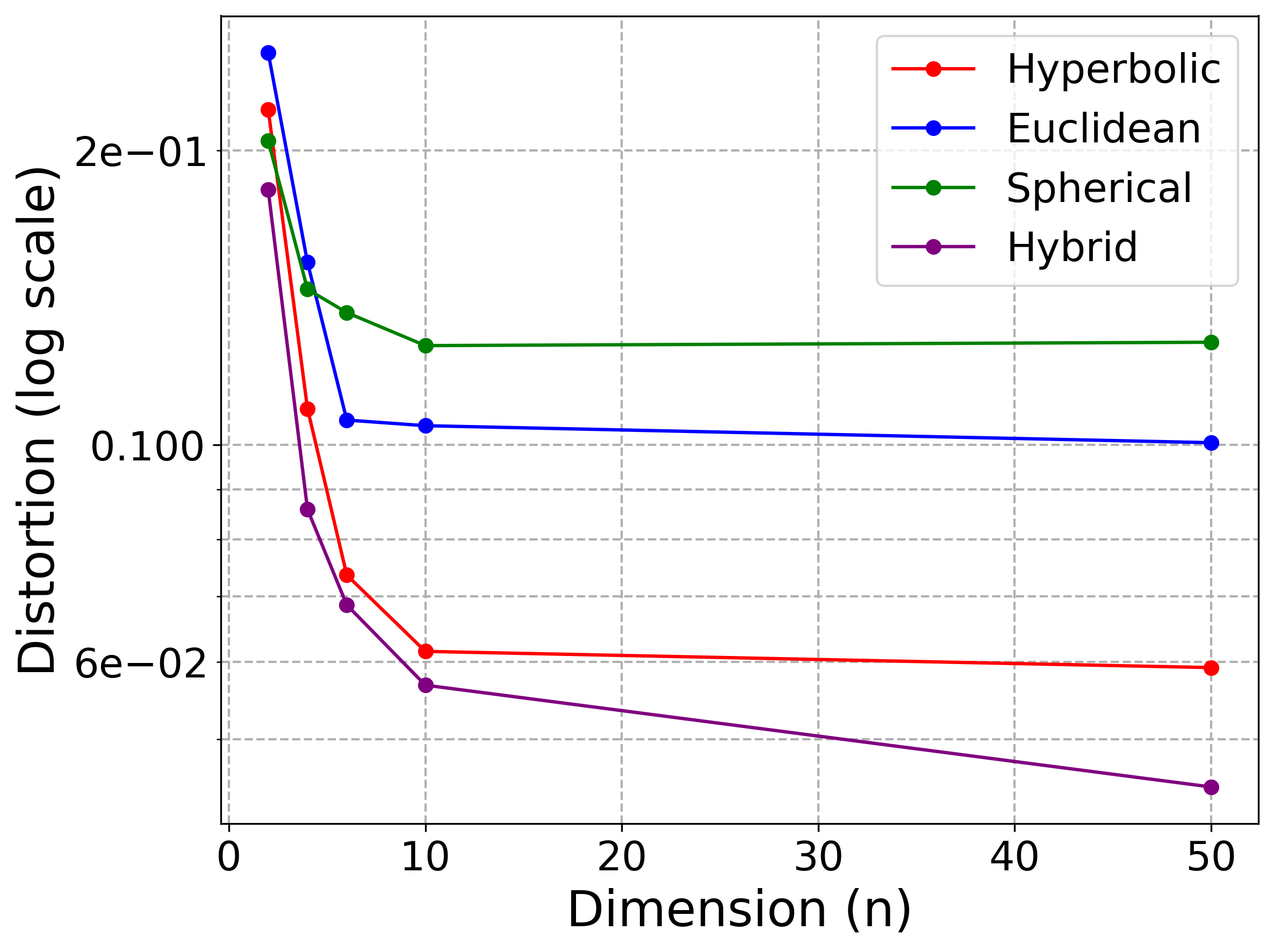}
        \caption{Ring of Trees}
        \label{fig:rt}
    \end{subfigure}
    
    \caption{Distortion of embedding a complete tree, cycle, and ring of tree into manifolds of different dimensions (log scale). Each graph has 96 nodes. Euclidean embeddings is shown in blue. In all cases, non-Euclidean geometry achieves significantly smaller distortion with significantly fewer dimensions. The distortion for Euclidean embeddings always plateaus, demonstrating that it is not suited for embeddings each structures regardless of its dimension.}
    \label{fig:dist_vs_dim_all}
\end{figure}
In this section we provide more plots of the distortion of embedding graphs into manifold of varying dimensions. The plots are shown in \cref{fig:dist_vs_dim_all}. In all cases, non-Euclidean geometry achieves significantly smaller distortion with significantly fewer dimensions, reflecting Takeaway 1 in \cref{sec:euclidean_limitation}. The distortion for Euclidean embeddings always plateaus, demonstrating that it is not suited for embeddings each structures regardless of its dimension. On the other hand, the distortion for non-Euclidean embeddings is still being reduced with increased dimensionality for 2 of the structures. This reflects Takeaway 2 in \cref{sec:euclidean_limitation}. 

\subsection{Dataset Details}
For the evaluation of token embedding distribution in LLMs, we incorporated a wide range of datasets, including a subset of the RedPajama dataset~\cite{weber2024redpajama} encompassing the arXiv, C4, Common Crawl, GitHub, Wikipedia, and StackExchange datasets; math reasoning datasets such as GSM8K~\cite{cobbe2021gsm8k}, MATH50K~\cite{hendrycksmath2021},MAWPS~\cite{koncel2016mawps}, and SVAMP~\cite{patel2021SVAMP}; and common sense reasoning datasets, including BoolQ, WinoGrande~\cite{sakaguchi2019winogrande}, and OpenBookQA~\cite{mihaylov2018openbookqa}.

\subsection{More Statistics}
\begin{figure*} 
    \centering
       \includegraphics[width=0.75\linewidth]{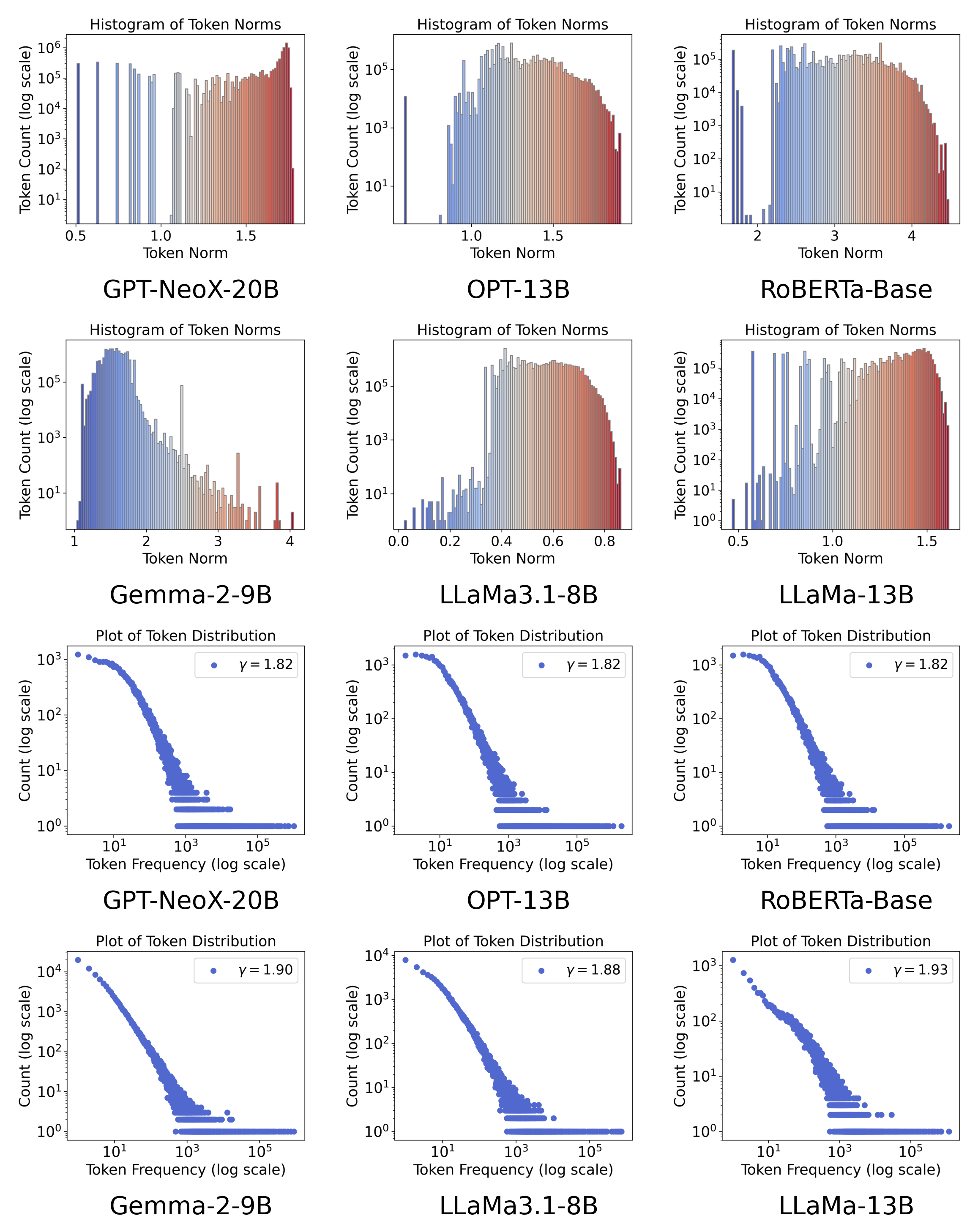}
   \caption{
   Token embeddings statistics for GPT-NeoX-20B, OPT-13B, RoBERT-Base, Gemma2-9B, LLaMa3.1-8B, and LLaMa-13B. The top 2 rows show distribution of the norm of the token embeddings and the bottom 2 rows show the distribution of the frequency of each token embedding. The token frequency distribution demonstrate scale-free property with power law decay, whereas the token norm show rapid decreases in token count for higher normed tokens at the right tail.}.
   \label{freq_norm_full}
\end{figure*}

In \cref{freq_norm_full} we show the statistics for token embeddings for more LLMs, including  GPT-NeoX-20B~\cite{gpt-neox-20b}, OPT-13B~\cite{zhang2022opt}, RoBERT-Base~\cite{liu2019roberta}, Gemma2-9B~\cite{gemma2}, LLaMa3.1-8B~\cite{dubey2024llama3}, and LLaMa-13B~\cite{touvron2023llama}. The top 2 rows show distribution of the norm of the token embeddings and the bottom 2 rows show the distribution of the frequency of each token embedding. The token frequency distribution demonstrate scale-free property with power law decay, whereas the token norm show rapid decreases in token count for higher normed tokens at the right tail. However, still none of the Euclidean foundational models fully capture the underlying scale-free property of the distribution, with all of them having an initial increase in token count against token norm for small normed token embeddings.

\begin{table*}[]
\centering
\caption{Hyperbolicity values $\delta$ for different metric spaces.}
\label{tab:hyperbolicity_metric}
\begin{tabular}{c|cccccc}
    \toprule
    & Sphere Space & Dense Graph & PubMed Graph & Poincare Space & Tree Graph \\
    \midrule
     $\delta$ & $0.99 \pm 0.01$ & $0.62 \pm 0.01$ & $0.40 \pm 0.04$ & $0.14 \pm 0.01$ & $0.0$ \\
    \bottomrule
\end{tabular}
\end{table*}

\section{$\delta$-Hyperbolicity Computation}
\label{appendix:delta_hyperbolicity}

Given any four points $a$, $b$, $c$, and $w$ in a metric space, the Gromov product $[a, c]_w$ at $w$ is bounded below by the minimum of the Gromov products $[a, b]_w$ and $[b, c]_w$, minus a slack term $\delta$:
\begin{equation}
[a, c]_w \geq \min([a, b]_w, [b, c]_w) - \delta.
\end{equation}
The Gromov product between $a$ and $b$ with respect to $w$ is defined as:
\begin{equation}
[a, b]_w = \frac{1}{2} \left( d(a, w) + d(b, w) - d(a, b) \right).
\end{equation}

A metric space $X$ is said to be $\delta$-hyperbolic if this inequality holds for all choices of $a$, $b$, $c$, and $w$. In geodesic metric spaces, $\delta$-hyperbolicity implies that geodesic triangles satisfy the $\delta$-slim property, meaning that any point on one side of a geodesic triangle is at most a distance of $\delta$ from some point on one of the other two sides.

In an exact tree metric, where the sides of any triangle intersect at a single point, the hyperbolicity constant $\delta$ is zero. This follows from the fact that the four-point condition holds as an equality for all points in the space.

\section{Foundational Operations for Pretraining Non-Euclidean Foundation Models}

\begin{table*}[t]
\centering
\caption{Geometric Foundation Model Operations: Euclidean vs. Manifold Formulations. $\operatorname{PT}_\mathcal{M}$:Parallel transport preserving vector properties during translation; $\exp_{\mu_\mathcal{M}}$: Exponential map from tangent space at Fréchet mean $\mu_\mathcal{M}$; $\log_{\mu_\mathcal{M}}$: Inverse exponential map projecting to tangent space;  $\operatorname{Ret}$: Retraction mapping for parameter updates; $\text{Proj}$: Tangent space projection operator}
\label{tab:formula}
\resizebox{0.95 \textwidth}{!}{
\begin{threeparttable}
\begin{tabular}{@{}lcc@{}}
\toprule
\centering
\textbf{Operation} & \textbf{Euclidean Space} & \textbf{Manifold Space} \\
\midrule
Curvature (K) & $K = 0$ & $K \in \mathbb{R}$ \\
\midrule
Attention Score & 
$\alpha_{q k}=\textrm{softmax}\left(\frac{q \cdot k^\top}{\sqrt{d_k}}\right)$ & 
$\alpha_{q k}^\mathcal{M}=\textrm{softmax}\left(\frac{-d_\mathcal{M}^2(q, k)}{\sqrt{d_k}}\right)$ \\
\midrule
Rotary PE & 
$Q_i^{\text{RoPE}}=Q_i \operatorname{Rot}(\mathbf{p}_i)$;\quad 
$K_i^{\text{RoPE}}=K_i \operatorname{Rot}(\mathbf{p}_i)$ & 
$Q_i^{\text{RoPE}_\mathcal{M}} = \operatorname{PT}_{\mathcal{M}}(Q_i, \mathbf{p}_i)$;\quad 
$K_i^{\text{RoPE}_\mathcal{M}} = \operatorname{PT}_{\mathcal{M}}(K_i, \mathbf{p}_i)$ \\
\midrule
Residual Connection & 
$x^{(l+1)}=x^{(l)}+f(x^{(l)})$ & 
$x^{(l+1)}=\exp_{x^{(l)}}(\lambda \cdot f(x^{(l)}))$ \\
\midrule
Layer Norm & 
$\hat{x}=\frac{x-\mu}{\sigma}$ & 
$\hat{x} = \exp_{\mu_\mathcal{M}}\left(\frac{\log_{\mu_\mathcal{M}}(x)}{\sigma_\mathcal{M}}\right)$ \\
\midrule
Cross-Entropy Loss & 
$\mathcal{L}=-\sum_t \log p_t$ & 
$\mathcal{L}=-\sum_t \log\frac{\exp(-d_\mathcal{M}(z_t, z^*))}{\sum_{t'}\exp(-d_\mathcal{M}^2(z_t, z_{t'}))}$ \\
\midrule
Optimization & 
$\theta_{t+1}=\theta_t-\eta \nabla_\theta J(\theta)$ & 
$\theta_{t+1} = \operatorname{Ret}_{\theta_t}\left(-\eta \text{Proj}_{T_{\theta_t}\mathcal{M}} \nabla J\right)$ \\
\midrule
FFN & 
$y = W_2\sigma(W_1x+b_1)+b_2$ & 
$y = \exp_{\mathbf{0}}\left(W_2\sigma(\log_{\mathbf{0}}(W_1 \otimes x \oplus b_1))\right)$ \\
\midrule
Attention Aggregation & 
$h = \sum_i \alpha_i v_i$ & 
$h = \operatorname{WeightedExpSum}(\{v_i\}, \{\alpha_i\})$ \\
\bottomrule
\end{tabular}
\end{threeparttable}  
} 
\end{table*}

\cref{tab:formula} systematically compares foundational operations in Euclidean space with their adaptations to non-Euclidean manifold spaces, highlighting critical geometric modifications required for pretraining curvature-aware foundation models. Below, we explain the key components and their mathematical formulations:

\textbf{Curvature (K).}  In Euclidean space, curvature is fixed at \( K = 0 \), reflecting flat geometry. In manifold spaces, curvature \( K \in \mathbb{R} \) is a learnable or estimated parameter that defines the intrinsic geometry (hyperbolic \( K < 0 \), spherical \( K > 0 \), or mixed). This value influences all subsequent operations, requiring dynamic adjustments to distance metrics and parameter updates. Curvature estimation methods (e.g., Ollivier-Ricci for graphs or learned embeddings for non-graph data) ensure geometric consistency across tasks.

\textbf{Attention Mechanism.}  
Euclidean attention computes similarity via dot products \( \alpha_{qk} = \textrm{softmax}\left(\frac{q \cdot k^\top}{\sqrt{d_k}}\right) \), while manifold attention replaces this with geodesic distance: \( \alpha_{qk}^\mathcal{M} = \textrm{softmax}\left(\frac{-d_\mathcal{M}^2(q, k)}{\sqrt{d_k}}\right) \). The negative squared distance prioritizes proximity on the manifold, preserving geometric relevance. Aggregation uses weighted Fréchet means (via exponential maps) or tangent space projections to combine features without violating curvature constraints.

\textbf{Positional Encoding (Rotary PE).}
Euclidean positional encodings apply rotation matrices \( \operatorname{Rot}(\mathbf{p}i) \) to query/key vectors. For manifolds, parallel transport \( \operatorname{PT}_\mathcal{M} \) replaces rotations, translating positional shifts along geodesics while preserving vector orientation relative to the manifold’s curvature. This ensures positional relationships respect intrinsic geometry.

\textbf{Residual Connections.}
Standard residuals \( x^{(l+1)} = x^{(l)} + f(x^{(l)}) \) are replaced by manifold equivalents: \( x^{(l+1)} = \exp_{x^{(l)}}(\lambda \cdot f(x^{(l)})) \). Here, the exponential map \( \exp \) projects tangent space updates \( f(x^{(l)}) \) onto the manifold, scaled by \( \lambda \), to preserve geometric stability across layers.

\textbf{Layer Normalization.}
Euclidean layer norm standardizes features via \( \hat{x} = \frac{x - \mu}{\sigma} \). On manifolds, operations occur in the tangent space at the Fréchet mean \( \mu_\mathcal{M} \): \( \hat{x} = \exp_{\mu_\mathcal{M}}\left(\frac{\log_{\mu_\mathcal{M}}(x)}{\sigma_\mathcal{M}}\right) \), where \( \log_{\mu_\mathcal{M}} \) maps points to the tangent space for normalization before reprojection.

\textbf{Cross-Entropy Loss.}
The manifold loss \( \mathcal{L} = -\sum_t \log\frac{\exp(-d_\mathcal{M}(z_t, z^*))}{\sum_{t'}\exp(-d_\mathcal{M}^2(z_t, z_{t'}))} \) replaces Euclidean dot products with geodesic distances, ensuring probabilities reflect the manifold’s geometry. This penalizes deviations in the curved space rather than in a flat embedding.

\textbf{Optimization.}
Euclidean SGD \( \theta_{t+1} = \theta_t - \eta \nabla_\theta J(\theta) \) is adapted via retractions \( \operatorname{Ret}_{\theta_t} \), which map gradient steps \( -\eta \text{Proj}_{T_{\theta_t}\mathcal{M}} \nabla J \) from the tangent space back to the manifold, ensuring updates respect curvature constraints.

\textbf{Feed-Forward Network (FFN).}
Manifold FFNs \( y = \exp_{\mathbf{0}}\left(W_2\sigma(\log_{\mathbf{0}}(W_1 \otimes x \oplus b_1))\right) \) use Möbius operations (\( \otimes, \oplus \)) for linear transformations and biases, followed by activation in the tangent space. The exponential map \( \exp_{\mathbf{0}} \) ensures outputs remain on the manifold.

\textbf{Attention Aggregation.}
Instead of weighted sums \( h = \sum_i \alpha_i v_i \), manifolds use \( \operatorname{WeightedExpSum} \), which computes Fréchet means of values \( v_i \) weighted by \( \alpha_i \), ensuring aggregated features lie on the manifold.

These adaptations collectively enable pretraining in non-Euclidean spaces by preserving geometric integrity. Operations like parallel transport, exponential/log maps, and retractions ensure compatibility with curvature, while specialized normalization and loss functions align learning dynamics with the manifold’s intrinsic structure. The table underscores the necessity of redefining core components—from attention to optimization—to build effective foundation models for hyperbolic and mixed-curvature geometries.

\end{document}